\documentclass{article} % For LaTeX2e
\usepackage{iclr2025_conference,times}

\usepackage{amsmath,amsfonts,bm}

\def\eqref#1{equation~\ref{#1}}
\def\1{\bm{1}}

\DeclareMathAlphabet{\mathsfit}{\encodingdefault}{\sfdefault}{m}{sl}
\SetMathAlphabet{\mathsfit}{bold}{\encodingdefault}{\sfdefault}{bx}{n}

\usepackage{url}

\usepackage[pagebackref=true,breaklinks=true,letterpaper=true,colorlinks, citecolor=darkerblue,bookmarks=false]{hyperref}

\usepackage{microtype}
\usepackage{graphicx}
\usepackage{subfigure}
\usepackage{booktabs} % for professional tables
\usepackage{amsmath}
\usepackage{amssymb}
\usepackage{mathtools}
\usepackage{amsthm}
\usepackage{listings}
\usepackage{multirow}
\usepackage{float}
\usepackage{enumitem}

\usepackage{color}
\usepackage{xcolor}
\usepackage{wrapfig}
\usepackage{colortbl}
\usepackage[makeroom]{cancel}
\usepackage{pifont}% http://ctan.org/pkg/pifont
\usepackage{rotating} % rotatebox

\definecolor{gray}{rgb}{0.5,0.5,0.5}
\definecolor{darkergreen}{RGB}{21, 152, 56}
\definecolor{darkerblue}{rgb}{0,0.08,0.45}
\definecolor{darkerred}{RGB}{220, 35, 120}
\definecolor{RoyalBlue}{RGB}{65,105,225}
\definecolor{YellowOrange}{RGB}{255,165,0}
\definecolor{gray94}{gray}{.94}
\definecolor{gray90}{gray}{.90}

\definecolor{plblue}{RGB}{100,59,237}

\newcolumntype{g}{>{\columncolor{gray94}}c} % column as gray
\newcommand{\rbox}[1]{\rotatebox{50}{#1}} % rotate cell

\newcommand{\cmark}{\ding{51}}%
\newcommand{\xmarkg}{\textcolor{gray}{\ding{55}}}%

\title{OpenMixup: Open Mixup Toolbox and Benchmark for Visual Representation Learning}

\author{
    Siyuan Li$^{1,2}$\thanks{Equal contribution.\ \ \ \ $^\dag$Corrsponding author.}\ \ \ %\textsuperscript{\rm 1}
    Zedong Wang$^{1*}$\ \ %\textsuperscript{\rm 1}
    Zicheng Liu$^{1,2}$\ \ %\textsuperscript{\rm 1}
    Juanxi Tian$^{1}$\ \ %\textsuperscript{\rm 1}
    Di Wu$^{1,2}$\\ %\textsuperscript{\rm 1}
    ~\textbf{Cheng Tan}$^{1,2}$\ \ %\textsuperscript{\rm 1}
    \textbf{Weiyang Jin}$^{1}$\ \ %\textsuperscript{\rm 1}
    \textbf{Stan Z. Li}$^{1\dag}$\\%\thanks{Corrsponding Author.}\\
    $^{1}$AI Lab, Research Center for Industries of the Future, Westlake University, Hangzhou, China\\
    $^{2}$Zhejiang University, Hangzhou, China\\
}

\iclrfinalcopy % Uncomment for camera-ready version, but NOT for submission.
\begin{document}

\maketitle

%%%%%%%%% ABSTRACT
\begin{abstract}

Mixup augmentation has emerged as a widely used technique for improving the generalization ability of deep neural networks (DNNs). However, the lack of standardized implementations and benchmarks has impeded recent progress, resulting in poor reproducibility, unfair comparisons, and conflicting insights. 
In this paper, we introduce OpenMixup, the \textit{first} mixup augmentation codebase and benchmark for visual representation learning. Specifically, we train 18 representative mixup baselines \textit{from scratch} and rigorously evaluate them across 11 image datasets of varying scales and granularity, ranging from fine-grained scenarios to complex non-iconic scenes.
We also open-source our modular codebase including a collection of popular vision backbones, optimization strategies, and analysis toolkits, which not only supports the benchmarking but enables broader mixup applications beyond classification, such as self-supervised learning and regression tasks.
%facilitating streamlined mixup method design, training, and evaluation for both researchers and practitioners in the community.
Through experiments and empirical analysis, we gain observations and insights on mixup performance-efficiency trade-offs, generalization, and optimization behaviors, and thereby identify preferred choices for different needs.
To the best of our knowledge, OpenMixup has facilitated several recent studies.
We believe this work can further advance reproducible mixup augmentation research and thereby lay a solid ground for future progress in the community.
The source code will be publicly available.
% The \href{https://github.com/Westlake-AI/openmixup}{source code} is publicly available.
% The source code for the benchmark is available at \url{https://github.com/Westlake-AI/openmixup}.

\end{abstract}

%%%%%%%%% BODY TEXT
% \begin{figure}[t]
\begin{wrapfigure}{r}{0.51\linewidth}
  \vspace{-5.0em}
  \begin{center}
\includegraphics[width=1.0\linewidth]{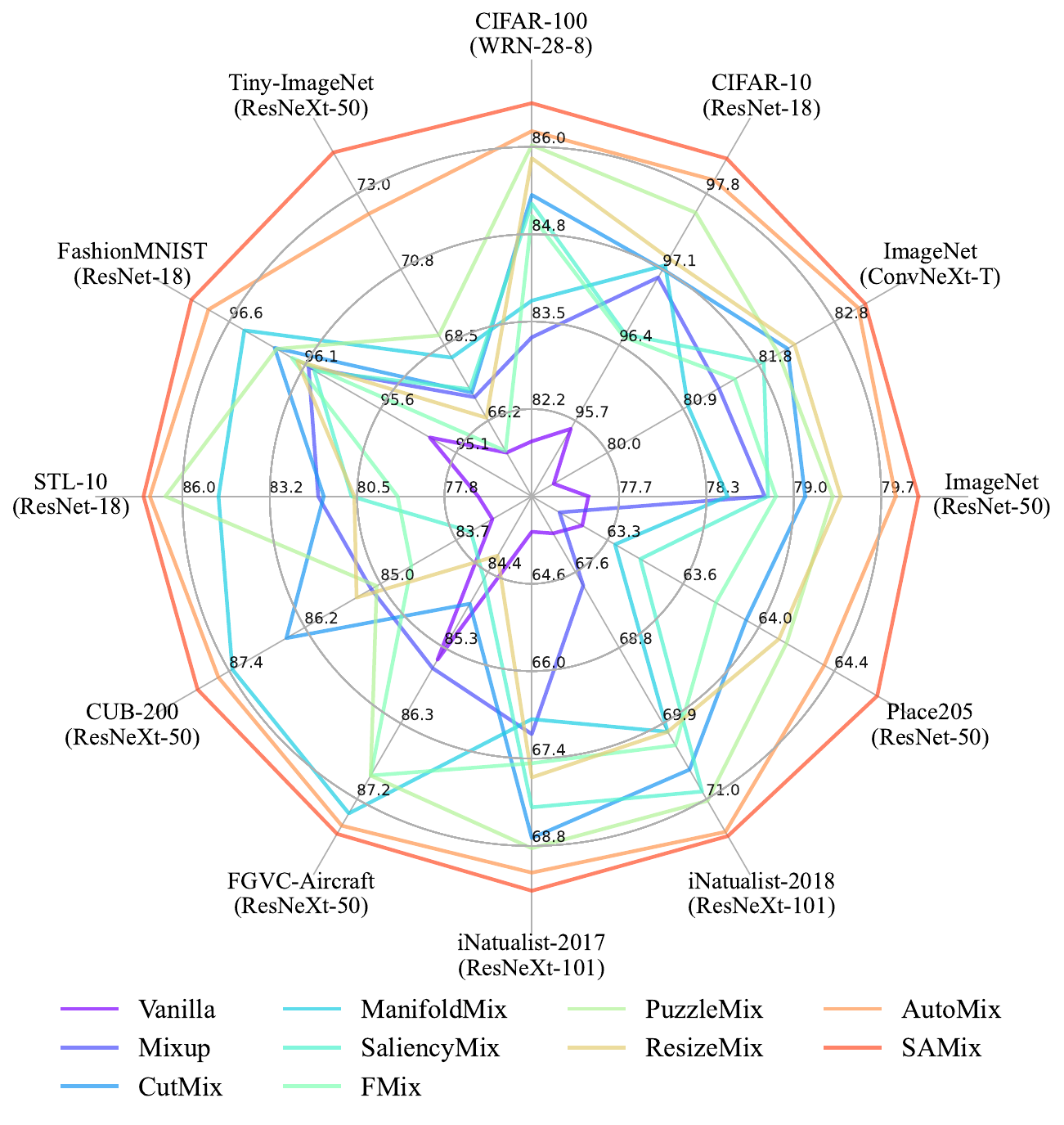}
  \end{center}
  \vspace{-1.75em}
  \caption{
  % Radar plot of top-1 accuracy for representative mixup baselines with CNN backbones on 11 image classification datasets.
  Radar plot of top-1 accuracy for representative mixup baselines on 11 classification datasets.
  }
  \label{fig:radar_dataset}
  \vspace{-1.5em}
\end{wrapfigure}
% \end{figure}

\section{Introduction}
\label{sec:introduction}
Data mixing, or mixup, has proven effective in enhancing the generalization ability of DNNs, with notable success in visual classification tasks.
The pioneering Mixup~\citep{Zhang2018mixupBE} proposes to generate mixed training examples through the convex combination of two input samples and their corresponding one-hot labels. 
By encouraging models to learn smoother decision boundaries, mixup effectively reduces overfitting and thus improves the overall performance.
%Such interpolation encourages models to learn smoother decision boundaries and thus increases data diversity. 
%
ManifoldMix \citep{Verma2019ManifoldMB} and PatchUp~\citep{2020patchup} extend this operation to the hidden space.
CutMix~\citep{Yun2019CutMixRS} presents an alternative approach, where an input rectangular region is randomly cut and pasted onto the target in the identical location.
Subsequent works~\citep{Harris2020FMixEM, Lee2020SmoothMixAS, Baek2021GridMixSR} have focused on designing more complex \textit{hand-crafted} policies to generate diverse and informative mixed samples, which can all be categorized as \textit{static} mixing methods.

\begin{figure*}[t]
    \vspace{-1.0em}
    \centering
    \includegraphics[width=1.0\textwidth]{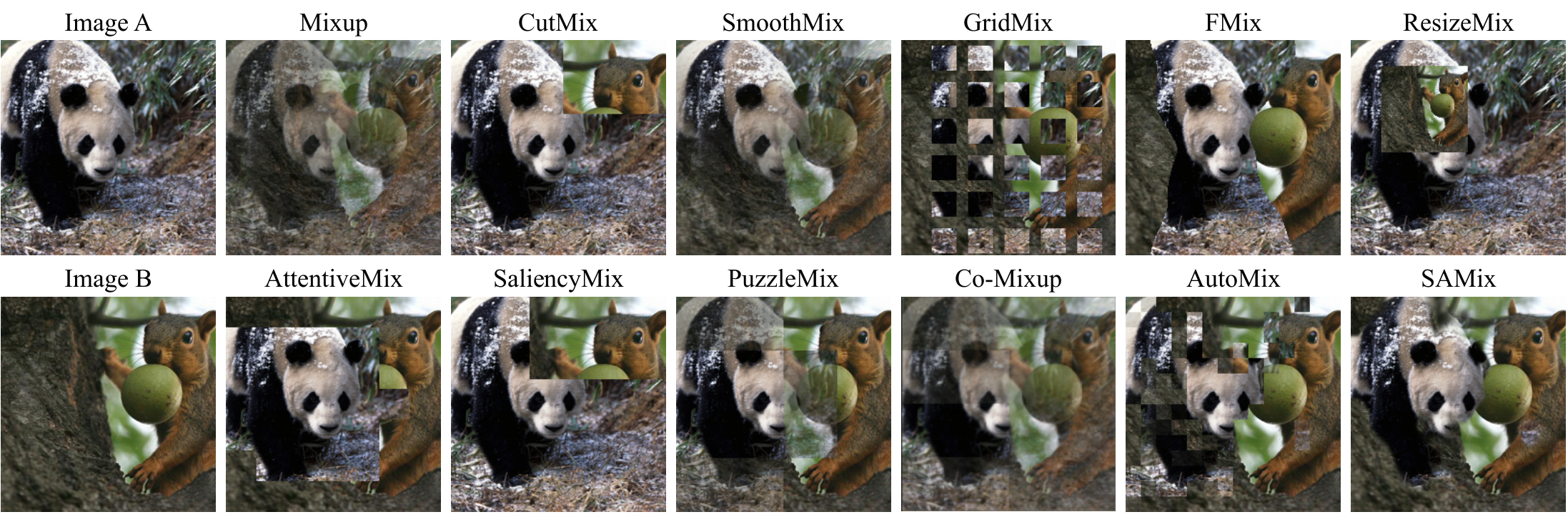}
    \vspace{-2.0em}
    \caption{
    Visualization of mixed samples from representative \textit{static} and \textit{dynamic} mixup augmentation methods on ImageNet-1K. We employ a mixing ratio of $\lambda=0.5$ for a comprehensive comparison. Note that mixed samples are more precisely in \textit{dynamic} mixing policies than these \textit{static} ones.
    }
    \label{fig:mixup}
    \vspace{-1.5em}
\end{figure*}

Despite efforts to incorporate saliency information into \textit{static} mixing framework~\citep{icassp2020Attentive, Uddin2021SaliencyMixAS, 2020resizemix}, they still struggle to ensure the inclusion of desired targets in the mixed samples, which may result in the issue of label mismatches. 
To address this problem, a new class of optimization-based methods, termed \textit{dynamic} mixing, has been proposed, as illustrated in the second row of Figure~\ref{fig:mixup}.
PuzzleMix~\citep{Kim2020PuzzleME} and Co-Mixup~\citep{Kim2021CoMixupSG} are two notable studies that leverage optimal transport to improve offline mask determination. 
More recently, TransMix~\citep{cvpr2022transmix}, TokenMix~\citep{eccv2022tokenmix}, MixPro~\citep{zhao2023mixpro}, and SMMix~\citep{iccv2023SMMix} are specifically tailored for Vision Transformers~\citep{iclr2021vit}.
The AutoMix series~\citep{Liu2021AutoMixUT, iclr2024adautomix} introduces a brand-new mixup learning paradigm, where mixed samples are computed by an online-optimizable generator in an end-to-end manner.
These emerging \textit{dynamic} approaches represent a promising avenue for generating semantically richer training samples that align with the underlying structure of input data.

\textbf{Why do we call for a mixup augmentation benchmark?}
While \textit{dynamic} methods have shown signs of surpassing the \textit{static} ones, their indirect optimization process incurs significant computational overhead, which limits their efficiency and applicability. Therefore, without a systematic understanding, it is uncertain if \textit{dynamic} mixup serves as the superior alternative in vision tasks. 
%it is inadvisable to naively discard the widely-used \textit{static} mixup without a systematic understanding.
Moreover, a thorough and standardized evaluation of different \textit{dynamic} methods is also missing in the community.
Benchmark is exactly the way to establish such an understanding, which plays a pivotal role in driving research progress by integrating an agreed-upon set of tasks, impartial comparisons, and assessment criteria.
To the best of our knowledge, however, there have been no such comprehensive benchmarks for mixup augmentation to facilitate unbiased comparisons and practical use in visual recognition.

\textbf{Why do we need an open-source mixup codebase?}
Notably, most existing mixup techniques are crafted with diverse settings, tricks, and implementations, each with its own coding style. 
This lack of standardization not only hinders user-friendly reproduction and deployment but impedes further development, thus imposing costly trial-and-error on practitioners to determine the most appropriate mixup strategy for their specific needs in real-world applications.
%However, there is currently no standardized mixup codebase for unified data pre-processing, mixup augmentation, backbone selection, model training, and evaluation.
Hence, it is essential to develop a unified mixup visual representation learning codebase for standardized data pre-processing, mixup development, network architecture selection, model training,  evaluation, and empirical analysis.

In this paper, we present OpenMixup, the \textit{first} comprehensive benchmark for mixup augmentation in vision tasks.
Unlike previous work~\citep{Naveed2021SurveyIM, survey2023mixing}, we train and evaluate 18 methods that represent the foremost strands on 11 diverse image datasets, as illustrated in Figure~\ref{fig:radar_dataset}.
We also open-source a standardized mixup codebase for visual representation learning, where the overall framework is built up with modular components for data pre-processing, mixup augmentation, network backbone selection, optimization, and evaluations. The codebase not only powers our benchmarking but supports broader relatively under-explored mixup applications beyond classification, such as semi-supervised learning~\citep{nips2019mixmatch}, self-supervised learning~\citep{nips2020mochi, shen2022unmix}, and dense prediction tasks~\citep{2017iccvmaskrcnn, 2020YOLOv4}.

Furthermore, insightful observations are obtained by incorporating multiple evaluation metrics and analysis toolkits in our codebase, including GPU memory usage (Figure~\ref{fig:trade_off}), loss landscape (Figure~\ref{fig:loss_landscape}), Power Law (PL) exponent alpha metrics (Figure~\ref{fig:alpha_norm}), robustness and calibration (Table~\ref{tab:cls_cifar100_vit_robust}), \textit{etc}.
% including GPU memory usage, loss landscape, robustness, and calibration, as illustrated in Section~\ref{sec:exp}, Figure~\ref{fig:trade_off}, Table~\ref{tab:cls_cifar100_vit}, Table~\ref{tab:cls_cifar100_vit_robust} and more. 
For instance, despite the key role \textit{static} mixing plays in today's deep learning systems, we surprisingly find that its generalizability over diverse datasets and backbones is significantly inferior to that of \textit{dynamic} algorithms.
By ranking the performance and efficiency trade-offs, we reveal that recent \textit{dynamic} methods have already outperformed the \textit{static} ones.
This may suggest a promising breakthrough for mixup augmentation, provided that the \textit{dynamic} computational overhead can be further reduced. 
Overall, we believe these insights can facilitate better evaluation and comparisons of mixup methods, enabling a systematic understanding and thus paving the way for further advancements.

%Nevertheless, we believe a comprehensive benchmark is essential and urgently needed for advancing the field. As such, we carried all these out from scratch. 
Since such a first-of-its benchmark can be rather time- and resource-consuming and most current advances have focused on and stemmed from visual classification tasks, we centralize our benchmarking scope on classification while extending it to broader mixup applications with transfer learning. Meanwhile, we have already supported these downstream tasks and datasets in our open-source codebase, allowing practitioners to customize their mixup algorithms, models, and training setups in these relatively under-explored scenarios. Our key contributions can be summarized as follows:

\begin{itemize}[leftmargin=2.0em]
\vspace{-0.25em}
    \item We introduce OpenMixup, the \textit{first} comprehensive benchmarking study for mixup augmentation, where 18 representative baselines are trained from scratch and rigorously evaluated on 11 visual classification datasets, ranging from non-iconic scenes to gray-scale, fine-grained, and long tail scenarios. By providing a standard testbed and a rich set of evaluation protocols, OpenMixup enables fair comparisons, thorough assessment, and analysis of different mixup strategies.
    %\vspace{-0.25em}
    \item 
    To support reproducible mixup research and user-friendly method deployment, we provide an open-source codebase for visual representation learning. The codebase incorporates standardized modules for data pre-processing, mixup augmentation, backbone selection, optimization policies, and distributed training functionalities. Beyond the benchmark itself, our OpenMixup codebase is readily extensible and has supported semi- and self-supervised learning and visual attribute regression tasks, which further enhances its utility and potential benefits to the community.
    %\vspace{-0.25em}
    \item 
    Observations and insights are obtained through extensive analysis. We investigate the generalization ability of all evaluated mixup baselines across diverse datasets and backbones, compare their GPU memory footprint and computational cost, visualize the loss landscape and PL exponent alpha metrics to understand optimization behavior, and evaluate robustness against input corruptions and calibration performance. Furthermore, we establish comprehensive rankings in terms of their performance and applicability (efficiency and versatility), offering clear method guidelines for specific requirements. These findings not only present a firm grasp of the current mixup augmentation landscape but shed light on promising avenues for future advancements.
\end{itemize}

\section{Background and Related Work}
\label{sec:background}

\subsection{Problem Definition}
\paragraph{Mixup Training.}
We first consider the general image classification tasks with $k$ different classes: given a finite set of $n$ image samples $X=[x_i]_{i=1}^{n}\in \mathbb{R}^{n\times W\times H\times C}$ and their corresponding ground-truth class labels $Y = [y_i]_{i=1}^{n}\in \mathbb{R}^{n\times k}$, encoded by a one-hot vector $y_i\in \mathbb{R}^{k}$. 
We attempt to seek the mapping from input data $x_i$ to its class label $y_i$ modeled through a deep neural network $f_{\theta}:x\longmapsto y$ with parameters $\theta$ by optimizing a classification loss $\ell(.)$, say the cross entropy (CE) loss,
\begin{equation}
    \ell_{CE}(f_{\theta}(x), y) = -y\log f_{\theta}(x).
    \label{eq:basic_cls}
    \vspace{-0.25em}
\end{equation}
Then we consider the mixup classification task: given a sample mixing function $h$, a label mixing function $g$, and a mixing ratio $\lambda$ sampled from $Beta(\alpha,\alpha)$ distribution, we can generate the mixed data $X_{mix}$ with $x_{mix}=h(x_i,x_j,\lambda)$ and the mixed label $Y_{mix}$ with $y_{mix}=g(y_i,y_j,\lambda)$, where $\alpha$ is a hyper-parameter. Similarly, we learn $f_{\theta}:x_{mix}\longmapsto y_{mix}$ by the mixup cross-entropy (MCE) loss,
\begin{equation}
    \ell_{MCE} = \lambda \ell_{CE}(f_\theta(x_{mix}), y_i) +
              (1-\lambda) \ell_{CE}(f_\theta(x_{mix}), y_j).
    \label{eq:mixup_cls}
    \vspace{-0.5em}
\end{equation}

\paragraph{Mixup Reformulation.}
Comparing Eq.~(\ref{eq:basic_cls}) and Eq.~(\ref{eq:mixup_cls}), the mixup training has the following features: 
(1) extra mixup policies, $g$ and $h$, are required to generate $X_{mix}$ and $Y_{mix}$. 
(2) the classification performance of $f_{\theta}$ depends on the generation policy of mixup. Naturally, we can split the mixup task into two complementary sub-tasks: 
(i) mixed sample generation and (ii) mixup classification (learning objective). 
Notice that the sub-task (i) is subordinate to (ii) because the final goal is to obtain a stronger classifier. 
Therefore, from this perspective, we regard the mixup generation as an auxiliary task for the classification task. 
Since $g$ is generally designed as a linear interpolation, i.e., $g(y_i,y_j,\lambda) = \lambda y_i + (1-\lambda)y_j$, $h$ becomes the key function to determine the performance of the model. 
Generalizing previous offline methods, we define a parametric mixup policy $h_{\phi}$ as the sub-task with another set of parameters $\phi$. 
% To differentiate the function of losses, $cls$ denotes the classification task and $gen$ denotes the generation task:
The final goal is to optimize $\ell_{MCE}$ given $\theta$ and $\phi$ as:
\vspace{-0.25em}
\begin{equation}
    \mathop{{\rm min}}\limits_{\theta,~\phi} \ell_{MCE} \Big (f_{\theta} \big (h_{\phi}(x_i,x_j,\lambda) \big), g(y_i,y_j,\lambda) \Big).
    \label{eq:emix_cls}
    \vspace{-0.5em}
\end{equation}

\subsection{Sample Mixing}
% \vspace{-0.25em}
Within the realm of visual classification, prior research has primarily concentrated on refining the sample mixing strategies rather than the label mixing ones. 
In this context, most sample mixing methods are categorized into two groups: \textit{static} policies and \textit{dynamic} policies, as presented in Table~\ref{tab:method}.

% \vspace{-0.5em}
\paragraph{Static Policies.}
The sample mixing procedure in all \textit{static} policies is conducted in a \textit{hand-crafted} manner. Mixup~\citep{Zhang2018mixupBE} first generates artificially mixed data through the convex combination of two selected input samples and their associated one-hot labels. 
ManifoldMix variants~\citep{Verma2019ManifoldMB, 2020patchup} extend the same technique to latent space for smoother feature mixing. Subsequently, CutMix~\citep{Yun2019CutMixRS} involves the random replacement of a certain rectangular region inside the input sample while concurrently employing Drop Patch throughout the mixing process.
Inspired by CutMix, several researchers in the community have explored the use of saliency information~\citep{Uddin2021SaliencyMixAS} to pilot mixing patches, while others have developed more complex \textit{hand-crafted} sample mixing strategies~\citep{Harris2020FMixEM, Baek2021GridMixSR}.

\begin{table*}[t]
    \centering
    \vspace{-1.5em}
    \setlength{\tabcolsep}{0.2mm}
    \caption{Overview of all supported vision Mixup augmentation methods in OpenMixup. Note that Mixup and CutMix in label mixing indicate mixing the labels of two samples by linear interpolation or computing cut squares. The \textit{Perf.}, \textit{App.}, and \textit{Overall} denote the performance, applicability, and overall rankings of all methods, which are derived from average rankings across baselines (view \ref{app:ranking}).}
     % \vspace{0.25em}
\resizebox{1.0\linewidth}{!}{
    \begin{tabular}{lccccccccc}
    \toprule
                        Method                                             & Category & Publication  & Sample Mixing                     & Label Mixing     & Extra Cost  & ViT only  & Perf. & App.  & Overall \\ \hline
\rowcolor[HTML]{FDF0E2} Mixup~\small{\citep{Zhang2018mixupBE}}             & Static   & ICLR'2018    & Hand-crafted Interpolation        & Mixup            & \xmarkg     & \xmarkg   & 15    & 1     & 10    \\
\rowcolor[HTML]{FDF0E2} CutMix~\small{\citep{Yun2019CutMixRS}}             & Static   & ICCV'2019    & Hand-crafted Cutting              & CutMix           & \xmarkg     & \xmarkg   & 13    & 1     & 8    \\
\rowcolor[HTML]{FDF0E2} DeiT~\small{(CutMix+Mixup)~\citep{icml2021deit}}   & Static   & ICML'2021    & CutMix+Mixup                      & CutMix+Mixup     & \xmarkg     & \xmarkg   &  7     & 1      & 3     \\
\rowcolor[HTML]{FDF0E2} SmoothMix~\small{\citep{Lee2020SmoothMixAS}}       & Static   & CVPRW'2020   & Hand-crafted Cutting              & CutMix           & \xmarkg     & \xmarkg   & 18    & 1     & 13   \\
\rowcolor[HTML]{FDF0E2} GridMix~\small{\citep{Baek2021GridMixSR}}          & Static   & PR'2021      & Hand-crafted Cutting              & CutMix           & \xmarkg     & \xmarkg   & 17    & 1     & 12   \\
\rowcolor[HTML]{FDF0E2} ResizeMix~\small{\citep{2020resizemix}}            & Static   & CVMJ'2023    & Hand-crafted Cutting              & CutMix           & \xmarkg     & \xmarkg   & 10     & 1     & 5    \\
\rowcolor[HTML]{FDF0E2} ManifoldMix~\small{\citep{Verma2019ManifoldMB}}    & Static   & ICML'2019    & Latent-space Mixup                & Mixup            & \xmarkg     & \xmarkg   & 14    & 1     & 9    \\
\rowcolor[HTML]{FDF0E2} FMix~\small{\citep{Harris2020FMixEM}}              & Static   & arXiv'2020   & Fourier-guided Cutting            & CutMix           & \xmarkg     & \xmarkg   & 16    & 1     & 11    \\
\rowcolor[HTML]{FDF0E2} AttentiveMix~\small{\citep{icassp2020Attentive}}   & Static   & ICASSP'2020  & Pretraining-guided Cutting        & CutMix           & \cmark      & \xmarkg   & 9     & 3     & 6    \\
\rowcolor[HTML]{FDF0E2} SaliencyMix~\small{\citep{Uddin2021SaliencyMixAS}} & Static   & ICLR'2021    & Saliency-guided Cutting           & CutMix           & \xmarkg     & \xmarkg   & 11     & 1     & 6    \\
\rowcolor[HTML]{E7ECE4} PuzzleMix~\small{\citep{Kim2020PuzzleME}}          & Dynamic  & ICML'2020    & Optimal-transported Cutting       & CutMix           & \cmark      & \xmarkg   & 8     & 4     & 6    \\
\rowcolor[HTML]{E7ECE4} AlignMix~\small{\citep{2021alignmix}}              & Dynamic  & CVPR'2022    & Optimal-transported Interpolation & CutMix           & \cmark      & \xmarkg   & 12    & 2     & 8    \\
\rowcolor[HTML]{E7ECE4} AutoMix~\small{\citep{Liu2021AutoMixUT}}           & Dynamic  & ECCV'2022    & End-to-end-learned Cutting        & CutMix           & \cmark      & \xmarkg   & 3     & 6     & 4    \\
\rowcolor[HTML]{E7ECE4} SAMix~\small{\citep{Li2021BoostingDV}}             & Dynamic  & arXiv'2021   & End-to-end-learned Cutting        & CutMix           & \cmark      & \xmarkg   & 1     & 5     & 1    \\
\rowcolor[HTML]{E7ECE4} AdAutoMix~\small{\citep{iclr2024adautomix}}        & Dynamic  & ICLR'2024    & End-to-end-learned Cutting        & CutMix           & \cmark      & \xmarkg   & 2      & 7      & 4     \\ \hline
\rowcolor[HTML]{CFEFFF} TransMix~\small{\citep{cvpr2022transmix}}          & Dynamic  & CVPR'2022    & CutMix+Mixup                      & Attention-guided & \xmarkg     & \cmark    & 5     & 8     & 7    \\
\rowcolor[HTML]{CFEFFF} SMMix~\small{\citep{iccv2023SMMix}}                & Dynamic  & ICCV'2023    & CutMix+Mixup                      & Attention-guided & \xmarkg     & \cmark    & 4     & 8     & 6    \\
\rowcolor[HTML]{CFEFFF} DecoupledMix~\small{\citep{2022decouplemix}}       & Static   & NeurIPS'2023 & Any Sample Mixing Policies        & DecoupledMix     & \xmarkg     & \xmarkg   & 6     & 1     & 2     \\
    \bottomrule
    \end{tabular}
    }
    \label{tab:method}
    \vspace{-1.5em}
\end{table*}

\vspace{-0.5em}
\paragraph{Dynamic Policies.}
In contrast to \textit{static} mixing, \textit{dynamic} strategies are proposed to incorporate sample mixing into an adaptive optimization-based framework.
PuzzleMix variants~\citep{Kim2020PuzzleME, Kim2021CoMixupSG} introduce combinatorial optimization-based mixing policies in accordance with saliency maximization. 
SuperMix variants~\citep{cvpr2021supermix, icassp2020Attentive} utilize pre-trained teacher models to compute smooth and optimized samples. 
Distinctively, AutoMix variants~\citep{Liu2021AutoMixUT, Li2021BoostingDV} reformulate the overall framework of sample mixing into an \textit{online-optimizable} fashion where the model learns to generate the mixed samples in an end-to-end manner.
% More recently, some researchers have focused on label mixing algorithms and generating mixed labels based on attention maps or a decouple strategy \citep{2022decouplemix}, which achieves superior performances compared to optimizable methods. 

\vspace{-0.25em}
\subsection{Label Mixing}
\vspace{-0.25em}
Mixup~\citep{Zhang2018mixupBE} and CutMix~\citep{Yun2019CutMixRS} are two widely-recognized label mixing techniques, both of which are \textit{static}. 
Recently, there has been a notable emphasis among researchers on advancing label mixing approaches, which attain more favorable performance upon certain sample mixing policies. 
Based on Transformers, TransMix variants~\citep{cvpr2022transmix, eccv2022tokenmix, nips2022TokenMixup, iccv2023SMMix} are proposed to utilize class tokens and attention maps to adjust the mixing ratio. 
A decoupled mixup objective \citep{2022decouplemix} is introduced to force models to focus on those hard mixed samples, which can be plugged into different sample mixing policies.
Holistically, most existing studies strive for advanced sample mixing designs rather than label mixing. 
%Therefore, the topic of label mixing will not be extensively discussed in this paper.

\vspace{-0.25em}
\subsection{Other Applications}
\vspace{-0.25em}
Recently, mixup augmentation also has shown promise in more vision applications, such as semi-supervised learning~\citep{nips2019mixmatch, 2022decouplemix}, self-supervised pre-training \citep{nips2020mochi, shen2022unmix}, and visual attribute regression~\citep{wu2021align, 2020YOLOv4}. Although these fields are not as extensively studied as classification, our OpenMixup codebase has been designed to support them by including the necessary task settings and datasets. Its modular and extensible architecture allows researchers and practitioners in the community to effortlessly adapt and extend their models to accommodate the specific requirements of these tasks, enabling them to quickly set up experiments without building the entire pipeline from scratch. Moreover, our codebase will be well-positioned to accelerate the development of future benchmarks, ultimately contributing to the advancement of mixup augmentation across a diversity of visual representation learning tasks.

\vspace{-0.25em}
\section{OpenMixup}
\label{sec:method}
\begin{figure*}[t!]
    \vspace{-0.5em}
    \centering
    \includegraphics[width=0.925\textwidth]{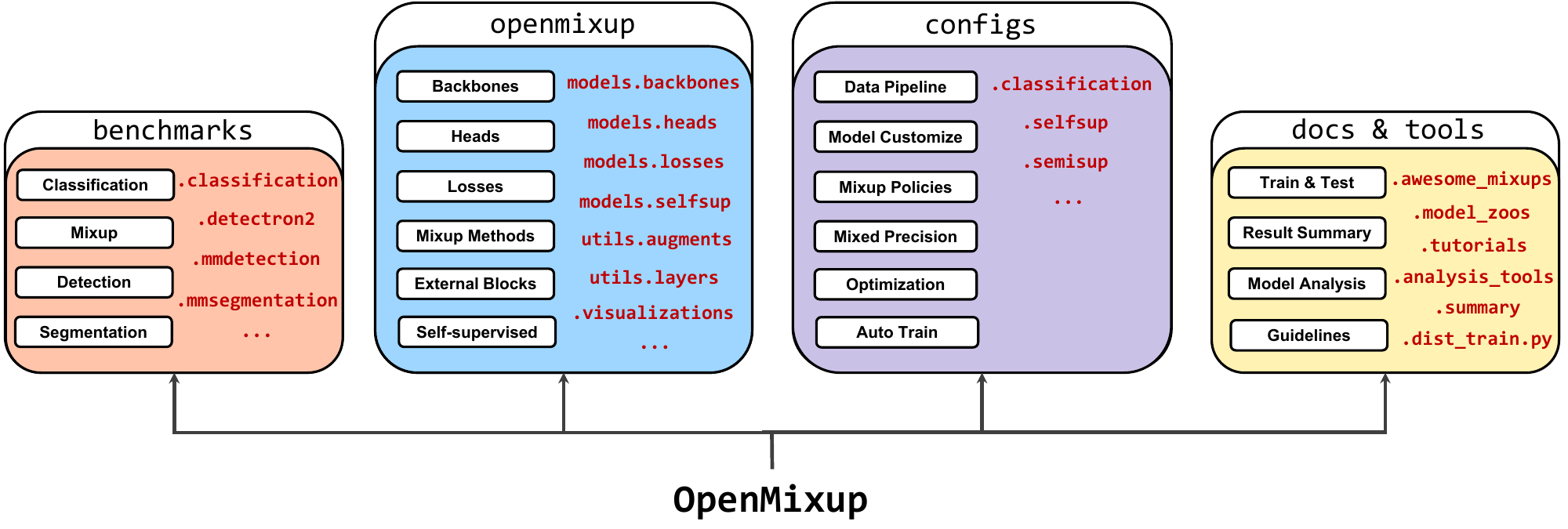}
    \vspace{-1.0em}
    \caption{Overview of codebase framework of OpenMixup. 
    (1) \texttt{benchmarks} provide benchmarking results and corresponding config files for mixup classification and transfer learning.
    (2) \texttt{openmixup} contains implementations of all supported methods.
    (3) \texttt{configs} is responsible for customizing setups of different mixup methods, networks, datasets, and training pipelines.
    (4) \texttt{docs \& tools} contains paper lists of popular mixup methods, user documentation, and useful tools.
    }
    \label{fig:overview}
    \vspace{-1.0em}
\end{figure*}

\vspace{-0.25em}
%In this section, we introduce our OpenMixup benchmark from four aspects: supported methods and tasks, evaluation metrics, and the experimental pipeline.
This section introduces our OpenMixup codebase framework and benchmark from four key aspects: supported methods and tasks, evaluation metrics, and experimental pipeline.
OpenMixup provides a unified framework implemented in PyTorch~\citep{nips2019pytorch} for mixup model design, training, and evaluation. The framework references MMClassification \citep{2020mmclassification} and follows the OpenMMLab coding style. 
We start with an overview of its composition.
As shown in Figure~\ref{fig:overview}, the whole training process here is fragmented into multiple components, including model architecture ({\color{purple}\texttt{.openmixup.models}}), data pre-processing ({\color{purple}\texttt{.openmixup.datasets}}), mixup policies ({\color{purple}\texttt{.openmixup.models.utils.augments}}), script tools ({\color{purple}\texttt{.tools}}) \textit{etc.} 
For instance, vision models are summarized into modular building blocks (\textit{e.g.}, backbone, neck, head \textit{etc.}) in {\color{purple}\texttt{.openmixup.models}}. 
This modular architecture enables practitioners to easily craft models by incorporating different components through configuration files in  {\color{purple}\texttt{.configs}}.
As such, users can readily customize their specified vision models and training strategies. 
In addition, benchmarking configuration ({\color{purple}\texttt{.benchmarks}}) and results ({\color{purple}\texttt{.tools.model\underline{ }zoos}}) are also provided in the codebase.
Additional benchmarking configurations and details are discussed below.

\subsection{Benchmarked Methods}
OpenMixup has implemented 17 representative mixup augmentation algorithms and 19 convolutional neural network and Transformer model architectures (gathered in {\color{purple}\texttt{.openmixup.models}}) across 12 diverse image datasets for supervised visual classification. 
We summarize these mixup methods in Table~\ref{tab:method}, along with their corresponding conference/journal, the types of employed sample, and label mixing policies, properties, and rankings.
For sample mixing, Mixup~\citep{Zhang2018mixupBE} and ManifoldMix~\citep{Verma2019ManifoldMB} perform \textit{hand-crafted} convex interpolation. CutMix~\citep{Yun2019CutMixRS}, SmoothMix~\citep{Lee2020SmoothMixAS}, GridMix~\citep{Baek2021GridMixSR} and ResizeMix~\citep{2020resizemix} implement \textit{hand-crafted} cutting policy. FMix~\citep{Harris2020FMixEM} utilizes Fourier-guided cutting. AttentiveMix~\citep{icassp2020Attentive} and SaliencyMix~\citep{Uddin2021SaliencyMixAS} apply pretraining-guided and saliency-guided cutting, respectively. 
Some \textit{dynamic} approaches like PuzzleMix~\citep{Kim2020PuzzleME} and AlignMix~\citep{2021alignmix} utilize optimal transport-based cutting and interpolation.
AutoMix~\citep{Liu2021AutoMixUT} and SAMix~\citep{Li2021BoostingDV} perform end-to-end online-optimizable cutting-based approaches.
As for the label mixing, most methods apply Mixup~\citep{Zhang2018mixupBE} or CutMix~\citep{Yun2019CutMixRS}, while the latest mixup methods for visual transformers (TransMix~\citep{cvpr2022transmix}, TokenMix~\citep{eccv2022tokenmix}, and SMMix~\citep{iccv2023SMMix}), as well as  DecoupledMix~\citep{2022decouplemix} exploit attention maps and a decoupled framework respectfully instead, which incorporate CutMix variants as its sample mixing strategy.
Such a wide scope of supported methods enables a comprehensive benchmarking analysis on visual classification.
% For self-supervised learning, 18 contrastive-based and auto-regressive algorithms have been supported. Additionally, 4 semi-supervised methods using mixup have been supported.

% \subsection{Supported Tasks}
\subsection{Benchmarking Tasks}
We provide detailed descriptions of the 12 open-source datasets as shown in Table~\ref{tab:datasets}. These datasets can be classified into four categories below:
\textbf{(1) Small-scale classification}: We conduct benchmarking studies on small-scale datasets to provide an accessible benchmarking reference. 
CIFAR-10/100~\citep{Krizhevsky2009Cifar} consists of 60,000 color images in 32$\times$32 resolutions. Tiny-ImageNet (Tiny)~\citep{Chrabaszcz2017TinyImageNet} and STL-10~\citep{icais2011stl10} are two re-scale versions of ImageNet-1K in the size of 64$\times$64 and 96$\times$96. FashionMNIST~\citep{Xiao2017FashionMNIST} is the advanced version of MNIST, which contains gray-scale images of clothing.
% \textbf{(1) Small-scale classification}: We conduct benchmarking study on small-scale datasets to provide \pl{a complete but accessible} benchmarking reference. 
% CIFAR-10/100~\citep{Krizhevsky2009Cifar} consists of 60,000 color images in 32$\times$32 resolutions, with 10 and 100 classes settings, respectively. Tiny-ImageNet (Tiny)~\citep{Chrabaszcz2017TinyImageNet}, as a re-scale version of ImageNet-1K, contains 10,000 training data and 10,000 validation data divided into 200 categories in size 64$\times$64.
%
\textbf{(2) Large-scale classification}: The large-scale dataset is employed to evaluate mixup algorithms against the most standardized procedure, which can also support the prevailing ViT architecture. ImageNet-1K (IN-1K)~\citep{Russakovsky2015ImageNetLS} is a well-known challenging dataset for image classification with 1000 classes.
% It contains 1.28 million training images with 469$\times$387 average resolutions grouped into 1000 classes. 
%
\textbf{(3) Fine-grained classification}: To investigate the effectiveness of mixup methods in complex inter-class relationships and long-tail scenarios, we conduct a comprehensive evaluation of fine-grained classification datasets, which can also be classified into small-scale and large-scale scenarios.
(i) \textit{Small-scale scenarios}: The datasets for small-scale fine-grained evaluation scenario are CUB-200-2011 (CUB)~\citep{Wah2011CUB2011} and FGVC-Aircraft (Aircraft)~\citep{Maji2013FineGrainedVC}, which contains a total of 200 wild bird species and 100 classes of airplanes.
% (i) \textit{Small-scale scenarios}: The datasets for small-scale fine-grained evaluation scenario are CUB-200-2011 (CUB)~\citep{Wah2011TheCB} and FGVC-Aircraft (Aircraft)~\citep{Maji2013FineGrainedVC}. CUB-200-2011 (CUB)~\citep{Wah2011TheCB} has a total of 200 wild bird species with 11,788 images for fine-grained classification. FGVC-Aircraft (Aircraft)~\citep{Maji2013FineGrainedVC} contains 10,000 images categorized into 100 classes of airplanes, 6,667 for training and 3,333 for testing. 
(ii) \textit{Large-scale scenarios}: The datasets for large-scale fine-grained evaluation scenarios are iNaturalist2017 (iNat2017)~\citep{cvpr2018inaturalist} and iNaturalist2018 (iNat2018)~\citep{cvpr2018inaturalist}, which contain 5,089 and 8,142 natural categories. Both the iNat2017 and iNat2018 own 7 major categories and are also long-tail datasets with scenic images (\textit{i.e.}, the fore-ground target is within large backgrounds). 
\textbf{(4) Scenic classification}: Scenic classification evaluations are also conducted to investigate the performance of different mixup augmentation methods in complex non-iconic scenarios on Places205~\citep{nips2014places205}.
% The dataset we used is Places205 \citep{nips2014places205}, which selects around 2,500,000 images as the training set and 41,000 images for validation from a total of 205 common scene classes.

\subsection{Evaluation Metrics and Tools}
We comprehensively evaluate the beneficial properties of mixup augmentation algorithms on the aforementioned vision tasks through the use of various metrics and visualization analysis tools in a rigorous manner. 
Overall, the evaluation methodologies can be classified into two distinct divisions, namely performance metric and empirical analysis. 
For the performance metrics, classification accuracy and robustness against corruption are two performance indicators examined.
As for empirical analysis, experiments on calibrations, CAM visualization, loss landscape, the plotting of training loss, and validation accuracy curves are conducted.
%We use them for specific tasks according to their characteristics.
The utilization of these approaches is contingent upon their distinct properties, enabling user-friendly deployment for designated purposes and demands.

\begin{table*}[t]
    \centering
    \vspace{-1.0em}
    \setlength{\tabcolsep}{0.5mm}
    \caption{The detailed information of supported visual classification datasets in OpenMixup.}
    % \vspace{-0.25em}
\resizebox{0.95\linewidth}{!}{
    \begin{tabular}{lcccccc}
    \toprule
Datasets                                         & Category                 & Source                                                                 & Classes & Resolution     & Train images & Test images \\ \hline
CIFAR-10~\citep{Krizhevsky2009Cifar}             & Iconic                   & \href{https://www.cs.toronto.edu/~kriz/cifar.html}{link}               & 10      & 32$\times$32   & 50,000       & 10,000      \\
CIFAR-100~\citep{Krizhevsky2009Cifar}            & Iconic                   & \href{https://www.cs.toronto.edu/~kriz/cifar.html}{link}               & 100     & 32$\times$32   & 50,000       & 10,000      \\
FashionMNIST~\citep{Xiao2017FashionMNIST}        & Gray-scale               & \href{https://github.com/zalandoresearch/fashion-mnist}{link}          & 10      & 28$\times$28   & 50,000       & 10,000      \\
STL-10~\citep{icais2011stl10}                    & Iconic                   & \href{https://cs.stanford.edu/~acoates/stl10/}{link}                   & 10      & 96$\times$96   & 50,00        & 8,000       \\
Tiny-ImageNet~\citep{Chrabaszcz2017TinyImageNet} & Iconic                   & \href{https://www.kaggle.com/c/tiny-imagenet}{link}                    & 200     & 64$\times$64   & 10,000       & 10,000      \\
ImageNet-1K~\citep{Russakovsky2015ImageNetLS}    & Iconic                   & \href{http://www.image-net.org/challenges/LSVRC/2012/}{link}           & 1000    & 469$\times$387 & 1,281,167    & 50,000      \\
CUB-200-2011~\citep{Wah2011CUB2011}              & Fine-grained             & \href{http://www.vision.caltech.edu/visipedia/CUB-200-2011.html}{link} & 200     & 224$\times$224 & 5,994        & 5,794       \\
FGVC-Aircraft~\citep{Maji2013FineGrainedVC}      & Fine-grained             & \href{https://www.robots.ox.ac.uk/~vgg/data/fgvc-aircraft/}{link}      & 100     & 224$\times$224 & 6,667        & 3,333       \\
iNaturalist2017~\cite{cvpr2018inaturalist}       & Fine-grained \& longtail & \href{https://github.com/visipedia/inat_comp/blob/master/2017}{link}   & 5089    & 224$\times$224 & 579,184      & 95,986      \\
iNaturalist2018~\cite{cvpr2018inaturalist}       & Fine-grained \& longtail & \href{https://github.com/visipedia/inat_comp/blob/master/2018}{link}   & 8142    & 224$\times$224 & 437,512      & 24,426      \\
Places205~\citep{nips2014places205}              & Scenic                   & \href{http://places.csail.mit.edu/downloadData.html}{link}             & 205     & 224$\times$224 & 2,448,873    & 41,000      \\
    \bottomrule
    \end{tabular}
    }
    \label{tab:datasets}
    \vspace{-1.0em}
\end{table*}

% metric
\vspace{-0.5em}
\paragraph{Performance Metric.}
\textbf{(1) Accuracy and training costs}: We adopt top-1 accuracy, total training hours, and GPU memory to evaluate all mixup methods' classification performance and training costs.
\textbf{(2) Robustness}: We evaluate the robustness against corruptions of the methods on CIFAR-100-C and ImageNet-C~\citep{Russakovsky2015ImageNetLS}, which is designed for evaluating the corruption robustness and provides 19 different corruptions, \textit{e.g.}, noise and blur \textit{etc.}
%
% \textbf{(3) Transfer to weakly supervised object localization}: We also evaluate the transferable abilities of the weakly supervised object localization (WSOL) task on CUB-200, which aims to localize objects of interest without bounding box supervision. 
\textbf{(3) Transferability to downstream tasks}: We evaluate the transferability of existing methods to object detection based on Faster R-CNN~\citep{ren2015faster} and Mask R-CNN~\citep{2017iccvmaskrcnn} on COCO~\textit{train2017}~\citep{eccv2014MSCOCO}, initializing with trained models on ImageNet. We also transfer these methods to semantic segmentation on ADE20K \citep{Zhou2018ADE20k}. Please refer to Appendix~\ref{app:transfer} for details.

% evaluation
\vspace{-0.5em}
\paragraph{Empirical Analysis.}
\textbf{(1) Calibrations}: To verify the calibration of existing methods, we evaluate them by the expected calibration error (ECE) on CIFAR-100~\citep{Krizhevsky2009Cifar}, \textit{i.e.}, the absolute discrepancy between accuracy and confidence. 
\textbf{(2) CAM visualization}: We utilize mixed sample visualization, a series of CAM variants~\citep{wacv2018GradCAMpp, ijcnn2020EigenCAM} (\textit{e.g.}, Grad-CAM~\citep{selvaraju2017gradcam}) to directly analyze the classification accuracy and especially the localization capabilities of mixup augmentation algorithms through top-1 top-2 accuracy predicted targets.
\textbf{(3) Loss landscape}: We apply loss landscape evaluation~\citep{nips2018VisualizingLoss} to further analyze the degree of loss smoothness of different mixup augmentation methods. 
\textbf{(4) Training loss and accuracy curve}: We plot the training losses and validation accuracy curves of various mixup methods to analyze the training stability, the ability to prevent over-fitting, and convergence speed.
\textbf{(5) Quality metric of learned weights}: Employing \texttt{WeightWatch}~\citep{NC2021weightwatcher}, we plot the Power Law (PL) exponent alpha metric of learned parameters with mixup algorithms to study their properties on different scenarios, \textit{e.g.,} acting as the regularizer to prevent overfitting or expanding more data as the augmentation technique to learn better representations.

% \begin{figure*}[t]  % Top
%     % \vspace{-0.5em}
% \centering
%     \subfigtopskip=-0.5pt
%     \subfigbottomskip=-0.5pt
%     \subfigcapskip=-4pt
%     \hspace{-0.5em}
%     \subfigure[\small{Performances on IN-1K}]{\label{fig:in1k_loss_r50}\includegraphics[width=0.54\linewidth,trim= 0 0 0 0,clip]{images/fig_radar_imagenet.pdf}}
%     \hspace{-0.5em}
%     \subfigure[\small{1D Loss landscapes}]{\label{fig:in1k_loss_r50}\includegraphics[width=0.462\linewidth,trim= 0 0 0 0,clip]{images/fig_R50_300ep_1d_loss_acc.pdf}}
% \vspace{-1.25em}
%     \caption{Visualization of various backbones performances and loss landscapes (1D) for multiple mixup methods on ImageNet-1K.
%     (a) Radar plots of top-1 accuracy using various network architectures.
%     (b) The training loss and validation top-1 accuracy are plotted with ResNet-50, showing that the dynamic Mixup methods achieve deeper and wider loss landscapes than the static variants.
%     % Using PyTorch-style training setting, models are trained 100 epochs with ResNet-18. The training loss and validation top-1 accuracy are plotted for compared mixup methods. We can find that the dynamic Mixup methods (\textit{e.g.,} PuzzleMix and AutoMix) achieve deepen and widen loss landscapes than the static variants (\textit{e.g.,} Mixup and CutMix).
%     }
%     \label{fig:loss_landscape}
%     \vspace{-0.5em}
% \end{figure*}

\subsection{Experimental Pipeline of OpenMixup Codebase}
OpenMixup provides a unified training pipeline that offers a consistent workflow across various computer vision tasks, as illustrated in Figure~\ref{fig:pipline}.
%With a unified training pipeline in OpenMixup, a comparable workflow is shared by different classification tasks, as illustrated in Figure~\ref{fig:pipline}.
Taking image classification as an example, we can outline the overall training process as follows.
(i) Data preparation: Users first select the appropriate dataset and pre-processing techniques from our supported data pipeline.
(ii) Model architecture: The {\color{purple}\texttt{openmixup.models}} module serves as a component library for building desired model architectures.
(iii) Configuration: Users can easily customize their experimental settings using Python configuration files under  {\color{purple}\texttt{.configs.classification}}. These files allow for the specification of datasets, mixup strategies, neural networks, and schedulers.
(iv) Execution: The {\color{purple}\texttt{.tools}} directory not only provides hardware support for distributed training but offers utility functionalities, such as feature visualization, model analysis, and result summarization, which can further facilitate empirical analysis.
% % arXiv
% Please refer to the online user documents at \url{https://openmixup.readthedocs.io/en/latest/} for more detailed guidelines (\textit{e.g.}, installation and getting started instructions), benchmarking results, awesome lists of related work, and more.
% Submit version
We also provide comprehensive online user documents, including detailed guidelines for installation and getting started instructions, all the benchmarking results, and awesome lists of related works in mixup augmentation, \textit{etc.}, which ensures that both researchers and practitioners in the community can effectively leverage our OpenMixup for their specific needs.

\section{Experiment and Analysis}
\label{sec:exp}
\subsection{Implementation Details}
\label{sec:setting}
We conduct essential benchmarking experiments of image classification on various scenarios with diverse evaluation metrics. For a fair comparison, grid search is performed for the shared hyper-parameter $\alpha\in \{0.1, 0.2, 0.5, 1, 2, 4\}$ of supported mixup variants while the rest of the hyper-parameters follow the original papers. Vanilla denotes the classification baseline without any mixup augmentations.
All experiments are conducted on Ubuntu workstations with Tesla V100 or NVIDIA A100 GPUs and report the \textit{mean} results of three trials. Appendix~\ref{app:benchmark} provides full visual classification results, Appendix~\ref{app:transfer} presents our transfer learning results for object detection and semantic segmentation, and Appendix~\ref{app:reproduction} conduct verification of the reproduction guarantee in OpenMixup.

\begin{figure*}[t]
\vspace{-2.25em}
\begin{minipage}{0.40\linewidth}
\centering
    \begin{table}[H]
    \caption{Top-1 accuracy (\%) on CIFAR-10/100 and Tiny-ImageNet (Tiny) based on ResNet (R), Wide-ResNet (WRN), and ResNeXt (RX) backbones.}
    % \vspace{-0.25em}
    \setlength{\tabcolsep}{1.3mm}
\resizebox{\linewidth}{!}{
\begin{tabular}{l|ccc}
    \toprule
                        Datasets      & CIFAR-10 & CIFAR-100 & Tiny   \\ \hline
                        Backbones     & R-18     & WRN-28-8  & RX-50  \\
                        Epochs        & 800 ep   & 800 ep    & 400 ep \\ \hline
\rowcolor[HTML]{FDF0E2} Vanilla       & 95.50    & 81.63     & 65.04  \\
\rowcolor[HTML]{FDF0E2} Mixup         & 96.62    & 82.82     & 66.36  \\
\rowcolor[HTML]{FDF0E2} CutMix        & 96.68    & 84.45     & 66.47  \\
\rowcolor[HTML]{FDF0E2} ManifoldMix   & 96.71    & 83.24     & 67.30  \\
\rowcolor[HTML]{FDF0E2} SmoothMix     & 96.17    & 82.09     & 68.61  \\
\rowcolor[HTML]{FDF0E2} AttentiveMix  & 96.63    & 84.34     & 67.42  \\
\rowcolor[HTML]{FDF0E2} SaliencyMix   & 96.20    & 84.35     & 66.55  \\
\rowcolor[HTML]{FDF0E2} FMix          & 96.18    & 84.21     & 65.08  \\
\rowcolor[HTML]{FDF0E2} GridMix       & 96.56    & 84.24     & 69.12  \\
\rowcolor[HTML]{FDF0E2} ResizeMix     & 96.76    & 84.87     & 65.87  \\
\rowcolor[HTML]{E7ECE4} PuzzleMix     & 97.10    & 85.02     & 67.83  \\
\rowcolor[HTML]{E7ECE4} Co-Mixup      & 97.15    & 85.05     & 68.02  \\
\rowcolor[HTML]{E7ECE4} AlignMix      & 97.05    & 84.87     & 68.74  \\
\rowcolor[HTML]{E7ECE4} AutoMix       & 97.34    & 85.18     & 70.72  \\
\rowcolor[HTML]{E7ECE4} SAMix         & 97.50    & \bf{85.50} & 72.18  \\
\rowcolor[HTML]{E7ECE4} AdAutoMix     & \bf{97.55} & 85.32   & \bf{72.89} \\
\rowcolor[HTML]{CFEFFF} Decoupled     & 96.95    & 84.88     & 67.46  \\
    \bottomrule
    \end{tabular}
    }
    \label{tab:cls_cifar}
\end{table}

\end{minipage}
~\begin{minipage}{0.59\linewidth}
\centering
    \begin{table}[H]
    \caption{Top-1 accuracy (\%) on ImageNet-1K using PyTorch-style, RSB A2/A3, and DeiT settings based on CNN and Transformer architectures, including ResNet (R), MobileNet.V2 (Mob.V2), DeiT-S, and Swin-T.}
    % \vspace{-0.25em}
    \setlength{\tabcolsep}{1.3mm}
\resizebox{\linewidth}{!}{
\begin{tabular}{l|ccccc}
    \toprule
                        Backbones     & R-50    & R-50   & Mob.V2 1x & DeiT-S & Swin-T \\ \hline
                        Epochs        & 100 ep  & 100 ep & 300 ep    & 300 ep & 300 ep \\
                        Settings      & PyTorch & RSB A3 & RSB A2    & DeiT   & DeiT   \\ \hline
\rowcolor[HTML]{FDF0E2} Vanilla       & 76.83   & 77.27  & 71.05     & 75.66  & 80.21  \\
\rowcolor[HTML]{FDF0E2} Mixup         & 77.12   & 77.66  & 72.78     & 77.72  & 81.01  \\
\rowcolor[HTML]{FDF0E2} CutMix        & 77.17   & 77.62  & 72.23     & 80.13  & 81.23  \\
\rowcolor[HTML]{FDF0E2} DeiT / RSB    & 77.35   & 78.08  & 72.87     & 79.80  & 81.20  \\
\rowcolor[HTML]{FDF0E2} ManifoldMix   & 77.01   & 77.78  & 72.34     & 78.03  & 81.15  \\
\rowcolor[HTML]{FDF0E2} AttentiveMix  & 77.28   & 77.46  & 70.30     & 80.32  & 81.29  \\
\rowcolor[HTML]{FDF0E2} SaliencyMix   & 77.14   & 77.93  & 72.07     & 79.88  & 81.37  \\
\rowcolor[HTML]{FDF0E2} FMix          & 77.19   & 77.76  & 72.79     & 80.45  & 81.47  \\
\rowcolor[HTML]{FDF0E2} ResizeMix     & 77.42   & 77.85  & 72.50     & 78.61  & 81.36  \\
\rowcolor[HTML]{E7ECE4} PuzzleMix     & 77.54   & 78.02  & 72.85     & 77.37  & 79.60  \\
\rowcolor[HTML]{E7ECE4} AutoMix       & 77.91   & 78.44  & 73.19     & 80.78  & 81.80  \\
\rowcolor[HTML]{E7ECE4} SAMix         & \bf{78.06} & \bf{78.64} & \bf{73.42} & 80.94  & \bf{81.87}  \\
\rowcolor[HTML]{E7ECE4} AdAutoMix     & 78.04   & 78.54  & -         & 80.81  & 81.75  \\
\rowcolor[HTML]{D1F5FF} TransMix      & -       & -      & -         & 80.68  & 81.80  \\
\rowcolor[HTML]{D1F5FF} SMMix         & -       & -      & -         & \bf{81.10}  & 81.80  \\
    \bottomrule
    \end{tabular}
    }
    \label{tab:cls_in1k}
\end{table}

\end{minipage}
\vspace{-1.0em}
\end{figure*}

\vspace{-0.5em}
\paragraph{Small-scale Benchmarks.}
We first provide standard mixup image classification benchmarks on five small datasets with two settings.
(a) The classical settings with the CIFAR version of ResNet variants~\citep{He2016DeepRL, xie2017aggregated}, \textit{i.e.}, replacing the $7\times 7$ convolution and MaxPooling by a $3\times 3$ convolution. We use $32\times 32$, $64\times 64$, and $28\times 28$ input resolutions for CIFAR-10/100, Tiny-ImageNet, and FashionMNIST, while using the normal ResNet for STL-10. We train vision models for multiple epochs from the stretch with SGD optimizer and a batch size of 100, as shown in Table~\ref{tab:cls_cifar} and Appendix~\ref{app:cifar_benchmark}.
(b) The modern training settings following DeiT~\citep{icml2021deit} on CIFAR-100, using $224\times 224$ and $32\times 32$ resolutions for Transformers (DeiT-S~\citep{icml2021deit} and Swin-T~\citep{iccv2021swin}) and ConvNeXt-T~\citep{2022convnet} as shown in Table~\ref{tab:cls_cifar100_vit}.
% As for the basic training data augmentation, we apply \texttt{RandomFlip} and \texttt{RandomCrop} with 4 pixels padding for 32$\times$32 resolutions. To fully explore the convergence speed, the properties of preventing over-fitting, and the performance limitations of various Mixup approaches, we train the models for 200/400/800/1200 epochs from stretch with SGD optimizer, a batch size of 100, and a basic learning rate of 0.1 adjusted by Cosine Scheduler~\citep{loshchilov2016sgdr} on CIFAR-10/100. Similarly, we adopt the basic augmentations of \texttt{RandomFlip} and \texttt{RandomResizedCrop} and optimize the models with a basic learning rate of 0.2 for 400 epochs with Cosine Scheduler. Table~\ref{tab:cls_cifar} shows parts of benchmark results, where Vanilla denotes the baseline using the raw CE loss.

\begin{figure*}[t]  % Top
    \vspace{-0.5em}
\centering
    \subfigtopskip=-0.5pt
    \subfigbottomskip=-0.5pt
    \subfigcapskip=-4pt
    \hspace{-0.5em}
    \subfigure[\small{DeiT-S on ImageNet-1K}]{\label{fig:trade_in1k_vit}\includegraphics[width=0.331\linewidth,trim= 0 0 0 0,clip]{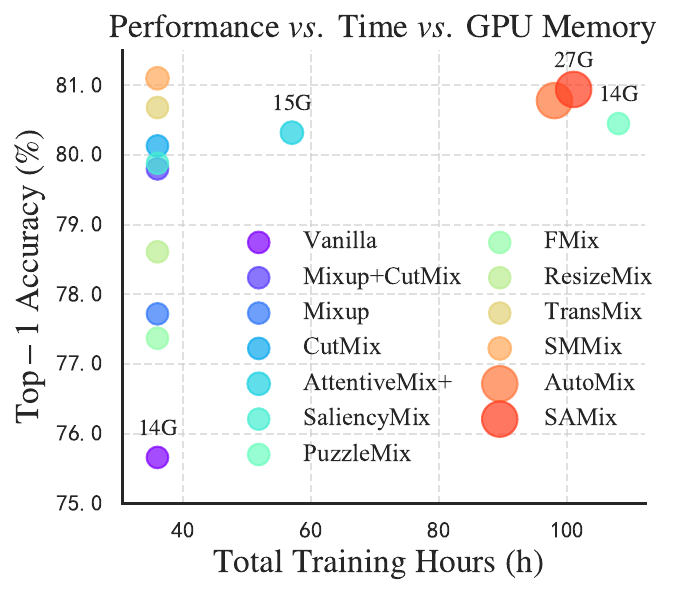}}
    \subfigure[\small{DeiT-S on CIFAR-100}]{\label{fig:trade_cifar_vit}\includegraphics[width=0.331\linewidth,trim= 0 0 0 0,clip]{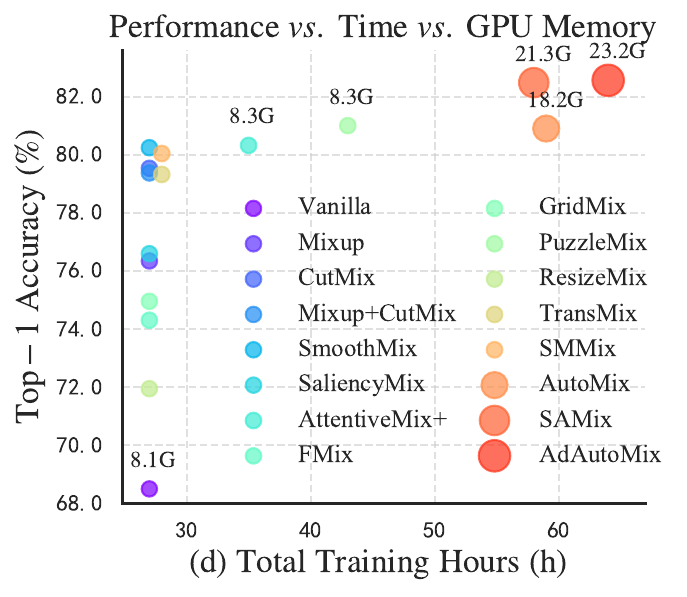}}
    \subfigure[\small{ConvNeXt-T on CIFAR-100}]{\label{fig:trade_cifar_cx}\includegraphics[width=0.332\linewidth,trim= 0 0 0 0,clip]{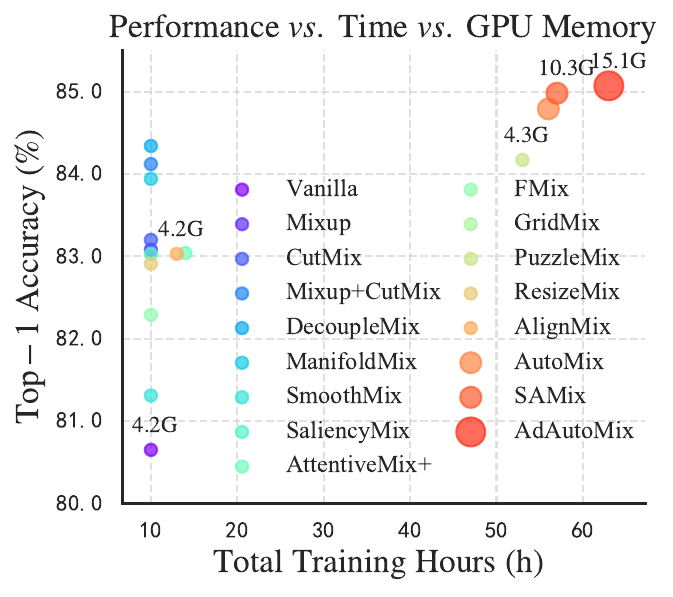}}
\vspace{-1.0em}
    \caption{Trade-off evaluation with respect to accuracy performance, total training time (hours), and GPU memory (G). The results in (a) are based on DeiT-S architecture on ImageNet-1K. The results in (b) and (c) are based on DeiT-S and ConvNeXt-T backbones on CIFAR-100, respectively.}
    \label{fig:trade_off}
    \vspace{-1.0em}
\end{figure*}

\vspace{-0.25em}
\paragraph{Standard ImageNet-1K Benchmarks.}
For visual augmentation and network architecture communities, ImageNet-1K is a well-known standard dataset. We support three popular training recipes: (a) PyTorch-style~\citep{He2016DeepRL} setting for classifical CNNs; (b) timm RSB A2/A3~\citep{Wightman2021rsb} settings; (c) DeiT~\citep{icml2021deit} setting for ViT-based models. Evaluation is performed on 224$\times$224 resolutions with \texttt{CenterCrop}. Popular network architectures are considered: ResNet~\citep{He2016DeepRL}, Wide-ResNet~\citep{bmvc2016wrn}, ResNeXt~\citep{xie2017aggregated}, MobileNet.V2~\citep{cvpr2018mobilenetv2}, EfficientNet~\citep{icml2019efficientnet}, DeiT~\citep{icml2021deit}, Swin~\citep{iccv2021swin}, ConvNeXt~\citep{2022convnet}, and MogaNet~\citep{Li2022MogaNet}. Refer to Appendix~\ref{app:implementation} for implementation details. In Table~\ref{tab:cls_in1k} and Table~\ref{tab:cls_in_torch}, we report the \textit{mean} performance of three trials where the \textit{median} of top-1 test accuracy in the last 10 epochs is recorded for each trial.
% The pre-trained checkpoints and logs will be made publicly available. More detailed benchmarking results and experiment ingredients are provided in the Appendix.

\vspace{-0.25em}
\paragraph{Benchmarks on Fine-grained and Scenic Scenarios.}
We further provide benchmarking results on three downstream classification scenarios in 224$\times$224 resolutions with ResNet backbone architectures: (a) Transfer learning on CUB-200 and FGVC-Aircraft. (b) Fine-grained classification on iNat2017 and iNat2018. (c) Scenic classification on Places205, as illustrated in Appendix~\ref{app:downstream} and Table~\ref{tab:cls_fgvc_place}.
% For (a), we use transfer learning settings on fine-grained datasets, using PyTorch official pre-trained models as initialization and training 200 epochs by SGD optimizer with the initial learning rate of 0.001, the weight decay of 0.0005, the batch size of 16, the same data augmentation as ImageNet-1K settings. For (b) and (c), we follow Pytorch-style ImageNet-1k settings mentioned above, training 100 epochs from stretch.

\begin{table*}[t]
    \centering
    \vspace{-0.5em}
    \caption{Rankings of various mixup augmentations as take-home messages for practical usage.}
    \setlength{\tabcolsep}{0.5mm}
    \setlength{\extrarowheight}{3pt}
    % \hspace{-1.5em}
    \vspace{2pt}
\resizebox{1.0\linewidth}{!}{
    \begin{tabular}{l|ccgccccccccccgcg}
    \toprule
              % & Mixup & CutMix & DeiT & Smooth & GridMix & ResizeMix & Manifold & FMix & Attentive & Saliency & PuzzleMix & AlignMix & AutoMix & SAMix & TransMix & SMMix \\ [3.6pt] \hline 
              & \rbox{Mixup} & \rbox{CutMix} & \rbox{DeiT} & \rbox{SmoothMix} & \rbox{GridMix} & \rbox{ResizeMix} & \rbox{ManifoldMix} & \rbox{FMix} & \rbox{AttentiveMix} & \rbox{SaliencyMix} & \rbox{PuzzleMix} & \rbox{AlignMix} & \rbox{AutoMix} & \rbox{SAMix} & \rbox{TransMix} & \rbox{SMMix} \\ \hline 
Performance   & 13    & 11     & 5    & 16     & 15      & 8         & 12       & 14   & 7         & 9        & 6         & 10       & 2       & 1     & 4        & 3     \\
Applicability & 1     & 1      & 1    & 1      & 1       & 1         & 1        & 1    & 3         & 1        & 4         & 2        & 7       & 6     & 5        & 5     \\ 
Overall       & 8     & 6      & 1    & 11     & 10      & 4         & 7        & 9    & 5         & 5        & 5         & 6        & 4       & 2     & 4        & 3     \\
    \bottomrule
    \end{tabular}
    }
    \label{tab:ranking}
    \vspace{-1.0em}
\end{table*}

% \paragraph{Empirical Observations and Insights.}
% \begin{enumerate}[(A)]
% \setlength\topsep{0.0em}
% \setlength\itemsep{0.10em}
% \setlength\leftmargin{0.5em}
\subsection{Observations and Insights}
\label{sec:observation}
Empirical analysis is conducted to gain insightful observations and a systematic understanding of the properties of different mixup augmentation techniques. Our key findings are summarized as follows:

\begin{figure*}[t]  % Top
    \vspace{-0.5em}
\centering
    \subfigtopskip=-0.5pt
    \subfigbottomskip=-0.5pt
    \subfigcapskip=-4pt
    % \hspace{-0.5em}
    \subfigure[DeiT-S on CIFAR-100]{\hspace{-0.25em}\label{fig:ep_acc_cifar_deit_s}\includegraphics[width=0.325\linewidth,trim= 0 0 0 0,clip]{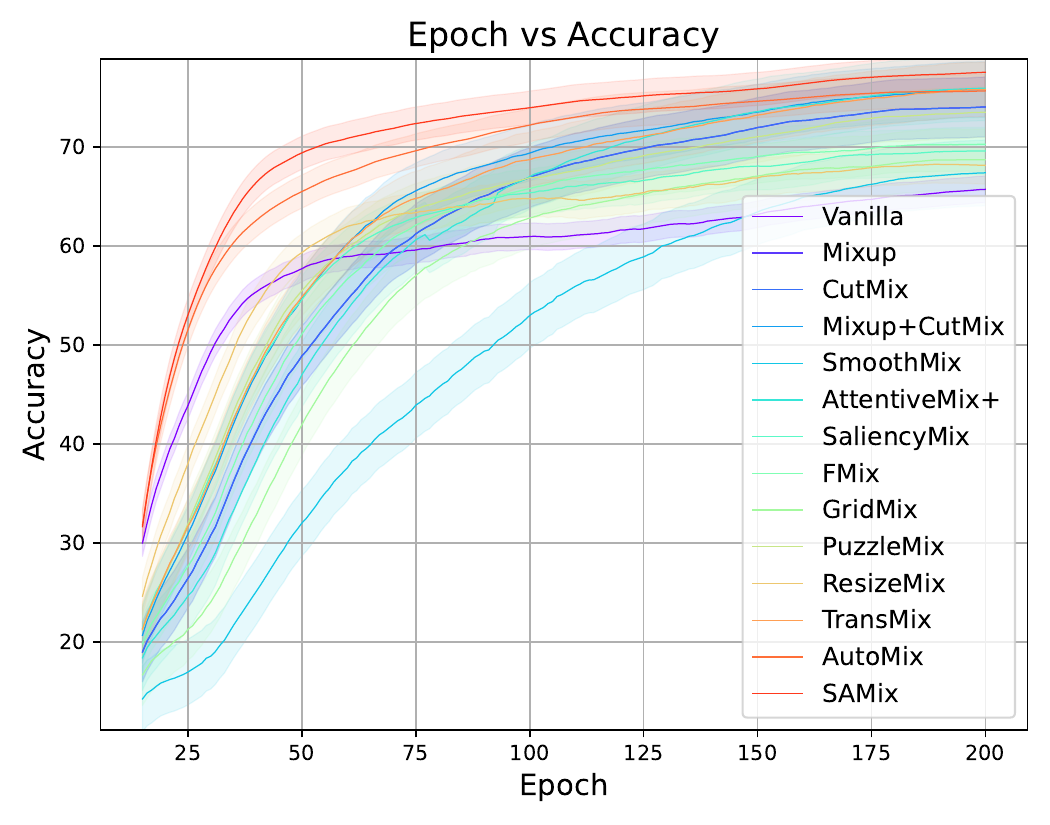}}
    \subfigure[Swin-T on CIFAR-100]{\hspace{-0.25em}\label{fig:ep_acc_cifar_swin_t}\includegraphics[width=0.325\linewidth,trim= 0 0 0 0,clip]{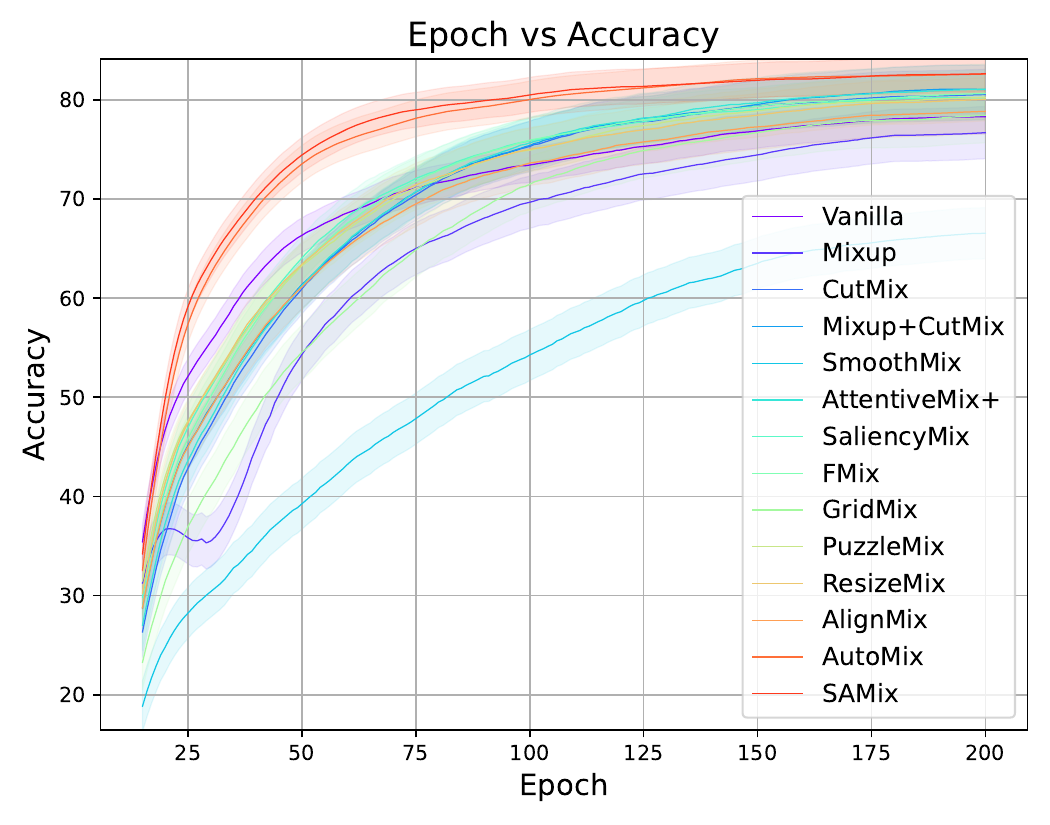}}
    \subfigure[ResNet-50 on ImageNet]{\hspace{-0.25em}\label{fig:loss_landscape}\includegraphics[width=0.355\linewidth,trim= 0 -10 0 0,clip]{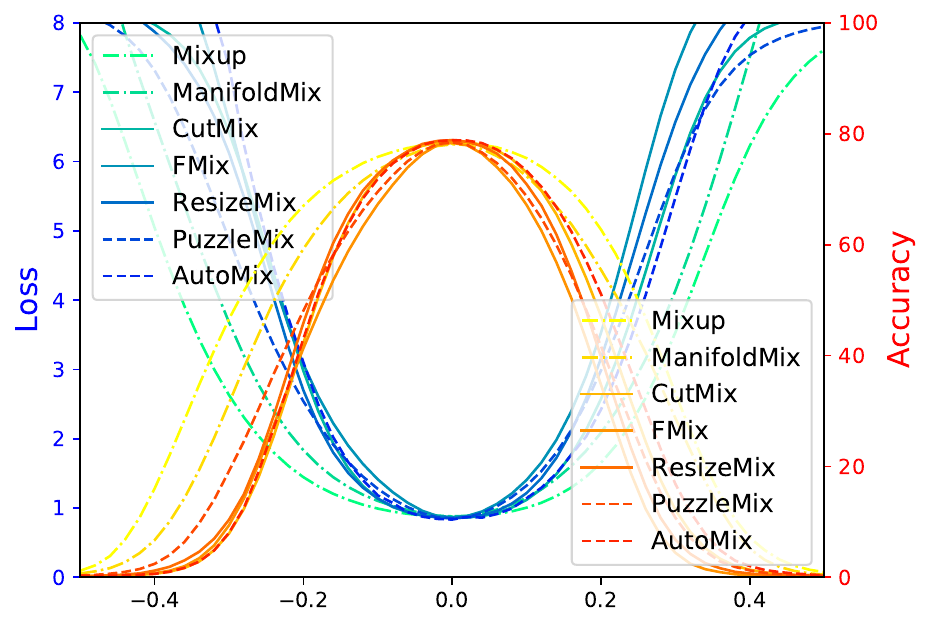}}
\vspace{-1.0em}
    \caption{(a)(b) Training epoch \textit{vs.} top-1 accuracy (\%) plots of different mixup methods on CIFAR-100 to analyze training stability and convergence speed. (c) 1-D loss landscapes for mixup methods with ResNet-50 (300 epochs) on ImageNet-1K. The results show that \textit{dynamic} approaches achieve deeper and wider loss landscapes than \textit{static} ones, which may indicate better optimization behavior.
    }
    \label{fig:epoch_vs_acc_landscape}
    \vspace{-0.5em}
\end{figure*}

\begin{figure*}[t]  % Top
    % \vspace{-0.5em}
\centering
    \subfigtopskip=-0.5pt
    \subfigbottomskip=-0.5pt
    \subfigcapskip=-4pt
    % \hspace{-0.5em}
    \subfigure[DeiT-S on CIFAR-100]{\hspace{-0.15em}\label{fig:alpha_cifar_deit}\includegraphics[width=0.33\linewidth,trim= 0 0 0 0,clip]{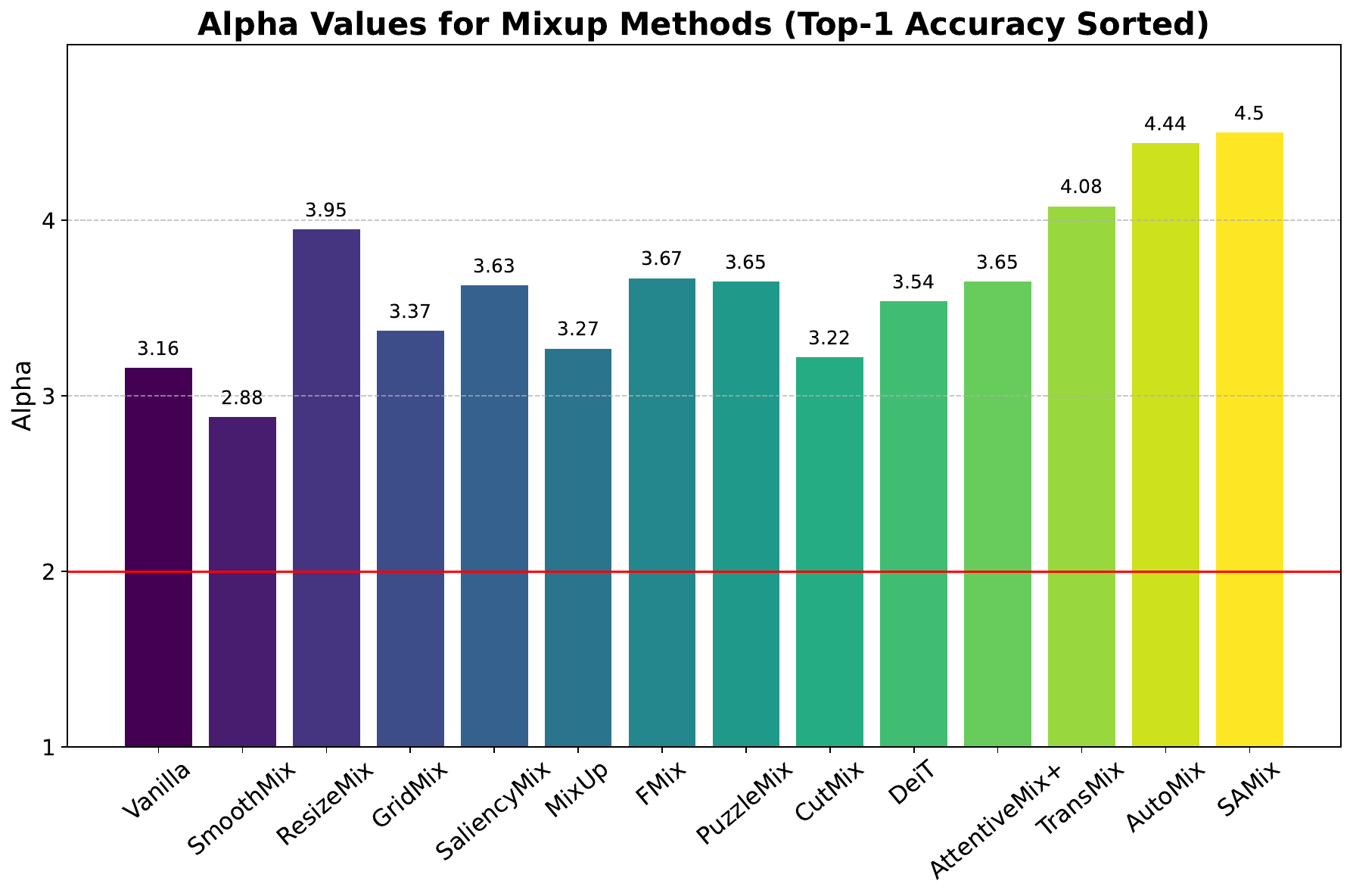}}
    \subfigure[Swin-T on CIFAR-100]{\hspace{-0.15em}\label{fig:alpha_cifar_swin}\includegraphics[width=0.33\linewidth,trim= 0 0 0 0,clip]{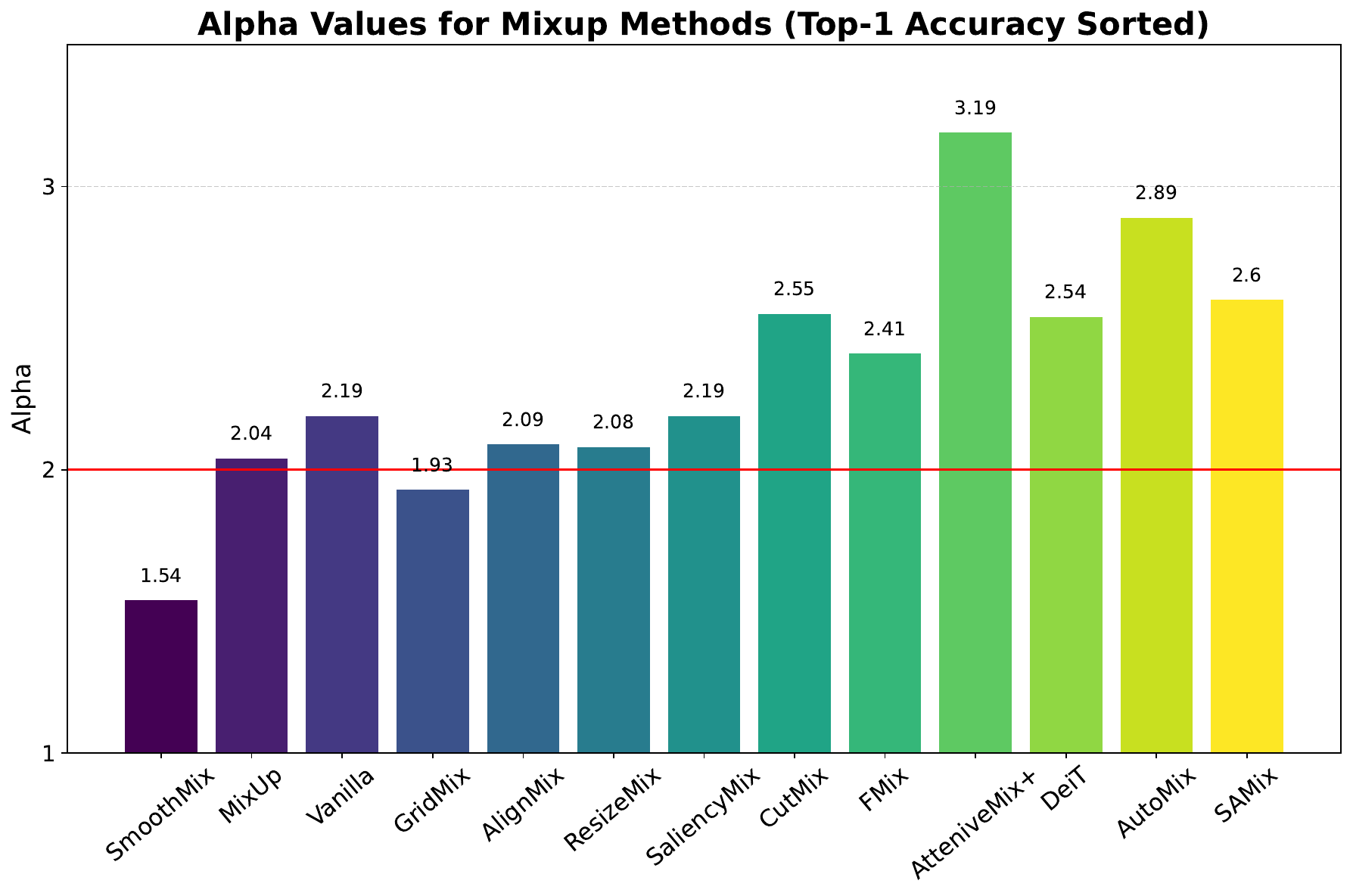}}
    \subfigure[DeiT-S on ImageNet-1K]{\hspace{-0.15em}\label{fig:alpha_in1k_deit}\includegraphics[width=0.33\linewidth,trim= 0 0 0 0,clip]{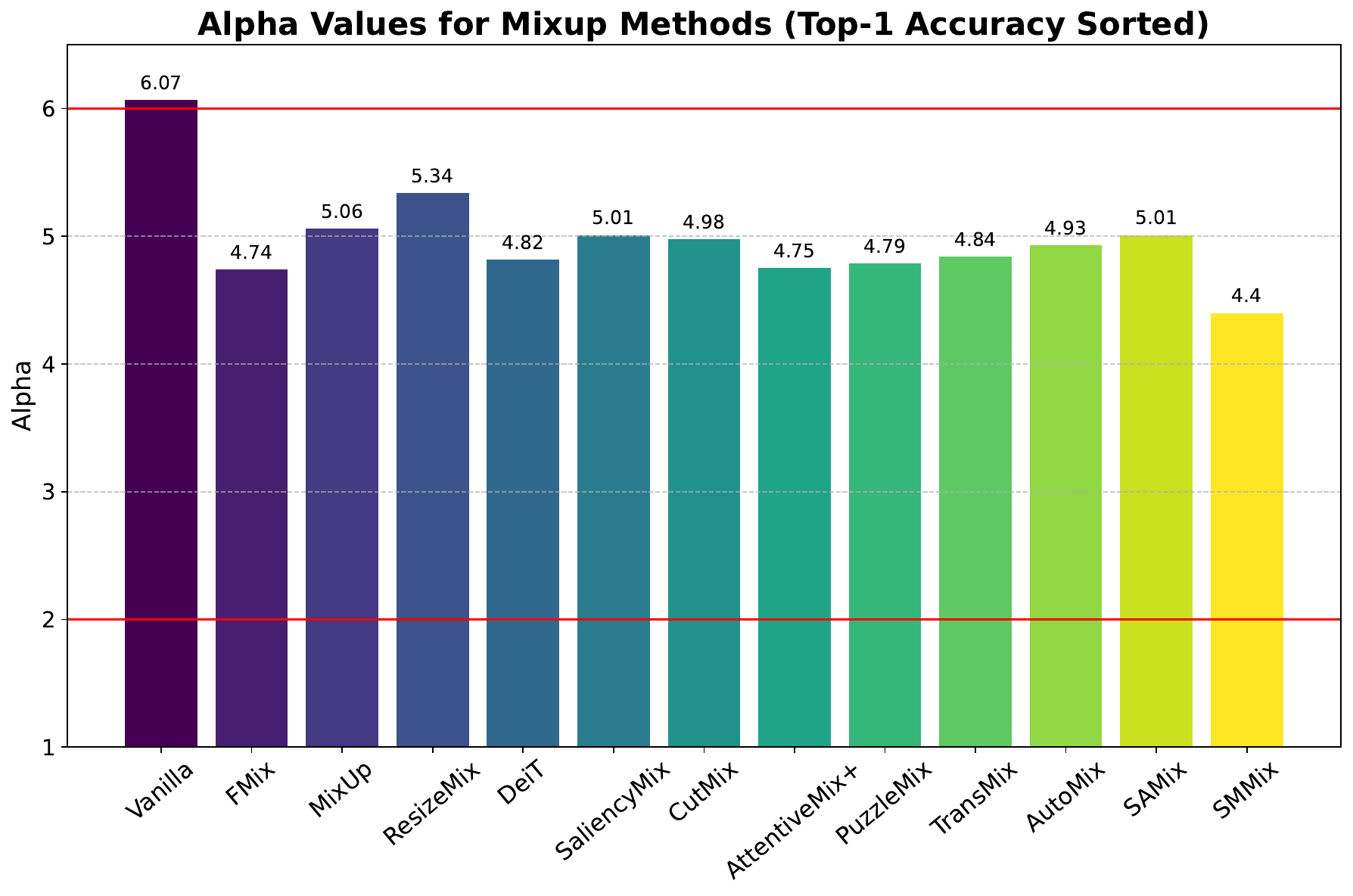}}
\vspace{-0.75em}
    \caption{Visualization of PL exponent alpha metrics~\citep{NC2021weightwatcher} of learned models by different mixup based on DeiT-S or Swin-T on (a)(b) CIFAR-100 and (c) ImageNet-1K. In each figure, the bars are sorted with the top-1 accuracy from left to right. Holistically, the alpha metric measures the fitting degree of the learned model to a certain task. A smaller alpha indicates better task fitting. Empirically, values less than 2 or larger than 6 run the risk of overfitting and underfitting. Therefore, this could serve as a favorable toolkit to evaluate the impact of different mixups on models.
    }
    \label{fig:alpha_norm}
    \vspace{-1.25em}
\end{figure*}

% \begin{wrapfigure}{r}{0.49\linewidth}
% % \begin{figure}[t]
%     \vspace{-1.25em}
%     \centering
%     \includegraphics[width=1.0\linewidth]{images/fig_R50_300ep_1d_loss_acc.pdf}
%     \vspace{-1.75em}
%     \caption{Visualization of 1D loss landscapes for representative mixup methods with ResNet-50 on ImageNet-1K. The validation accuracy is plotted, showing that dynamic methods achieve deeper and wider loss landscapes than static ones.
%     }
%     \label{fig:loss_landscape}
%     \vspace{-1.5em}
% % \end{figure}
% \end{wrapfigure}

\textbf{(A) Which mixup method should I choose?}
Integrating benchmarking results from various perspectives, we provide practical mixup rankings (detailed in Appendix~\ref{app:ranking}) as a take-home message for real-world applications, which regards performance, applicability, and overall capacity.
As shown in Table~\ref{tab:method}, as for the performance, the \textit{online-optimizable} SAMix and AutoMix stand out as the top two choices. SMMix and TransMix follow closely behind. However, regarding applicability that involves both the concerns of efficiency and versatility, \textit{hand-crafted} methods significantly outperform the learning-based ones.
Overall, the DeiT (Mixup+CutMix), SAMix, and SMMix are selected as the three most preferable mixup methods, each with its own emphasis. Table~\ref{tab:ranking} shows ranking results.

\textbf{(B) Generalization over datasets.} 
The intuitive performance radar chart presented in Figure~\ref{fig:radar_dataset}, combined with the trade-off results in Figure~\ref{fig:trade_off}, reveals that \textit{dynamic} mixup methods consistently yield better performance compared to \textit{static} ones, showcasing their impressive generalizability. 
However, \textit{dynamic} approaches necessitate meticulous tuning, which incurs considerable training costs. 
In contrast, \textit{static} mixup exhibits significant performance fluctuation across different datasets, indicating poor generalizability with application scenarios. For instance, Mixup and CutMix as the \textit{static} representatives perform even worse than the baseline on Place205 and FGVC-Aircraft, respectively.
Moreover, we analyze how mixup methods improve on different datasets in Figure~\ref{fig:alpha_norm} and Figure~\ref{fig:app_alpha_norm}. On small-scale datasets, mixup methods (\textit{dynamic} ones) tend to prevent the over-parameterized backbones (Vanilla or with some \textit{static} ones) from overfitting. On the contrary, mixup techniques are served as data augmentations to encourage the model to fit hard tasks on large-scale datasets.

\textbf{(C) Generalization over backbones.}
As shown in Figure~\ref{fig:trade_off} and Figure~\ref{fig:loss_landscape}, we provide extensive evaluations on ImageNet-1K based on different types of backbones and mixup methods. As a result, \textit{dynamic} mixup achieves better performance in general and shows more favorable generalizability against backbone selection compared to \textit{static} methods.
Noticeably, the \textit{online-optimizable} SAMix and AutoMix exhibit impressive generalization ability over different vision backbones, which potentially reveals the superiority of their online training framework compared to the others. 

\textbf{(D) Applicability.} 
Figure~\ref{fig:in1k_loss_r50} shows that ViT-specific methods (\textit{e.g.}, TransMix~\citep{cvpr2022transmix} and TokenMix~\citep{eccv2022tokenmix}) yield exceptional performance with DeiT-S and PVT-S yet exhibit intense sensitivity to different model scales (\textit{e.g.}, with PVT-T). Moreover, they are limited to ViTs, which largely restricts their applicability.
Surprisingly, \textit{static} Mixup~\citep{Zhang2018mixupBE} exhibits favorable applicability with new efficient networks like MogaNet~\citep{Li2022MogaNet}.
CutMix~\citep{Yun2019CutMixRS} fits well with popular backbones, such as modern CNNs (\textit{e.g.}, ConvNeXt and ResNeXt) and DeiT, which increases its applicability. As shown in Figure~\ref{fig:trade_off}, although AutoMix and SAMix are available in both CNNs and ViTs with consistent superiority, they have limitations in GPU memory and training time, which may limit their applicability in certain cases. This also provides a promising avenue for reducing the cost of well-performed online learnable mixup augmentation algorithms. 

\textbf{(E) Robustness \& Calibration.} 
We evaluate the robustness with accuracy on the corrupted version of CIFAR-100 and FGSM attack~\citep{iclr2015fgsm} and the prediction calibration. Table~\ref{tab:cls_cifar100_vit_robust} shows that all the benchmarked methods can improve model robustness against corruptions. However, only four recent \textit{dynamic} approaches exhibit improved robustness compared to the baseline with FGSM attacks. We thus hypothesize that the \textit{online-optimizable} mixup methods are robust against human interference, while the \textit{hand-crafted} ones adapt to natural disruptions like corruption but are susceptible to attacks.  Overall, AutoMix and SAMix achieve the optimal robustness and calibration results. For scenarios where these properties are required, practitioners can prioritize these methods.

\textbf{(F) Convergence \& Training Stability.}
As shown in Figure~\ref{fig:epoch_vs_acc_landscape}, wider bump curves indicate smoother loss landscapes (\textit{e.g.}, Mixup), while higher warm color bump tips are associated with better convergence and performance (\textit{e.g.}, AutoMix). 
Evidently, \textit{dynamic} mixup algorithms own better training stability and convergence than \textit{static} mixup in general while obtaining sharp loss landscapes. They are likely to improve performances through exploring hard mixup samples.
Nevertheless, the \textit{static} mixup variants with convex interpolation, especially vanilla Mixup, exhibit smoother loss landscape and stable training than some \textit{static} cutting-based methods.
Based on the observations, we assume this arises from its interpolation that prioritizes training stability but may lead to sub-optimal results.

\begin{figure*}[t]
    \vspace{-0.5em}
    \centering
    \hspace{-0.25em}
    \includegraphics[width=1.002\textwidth]{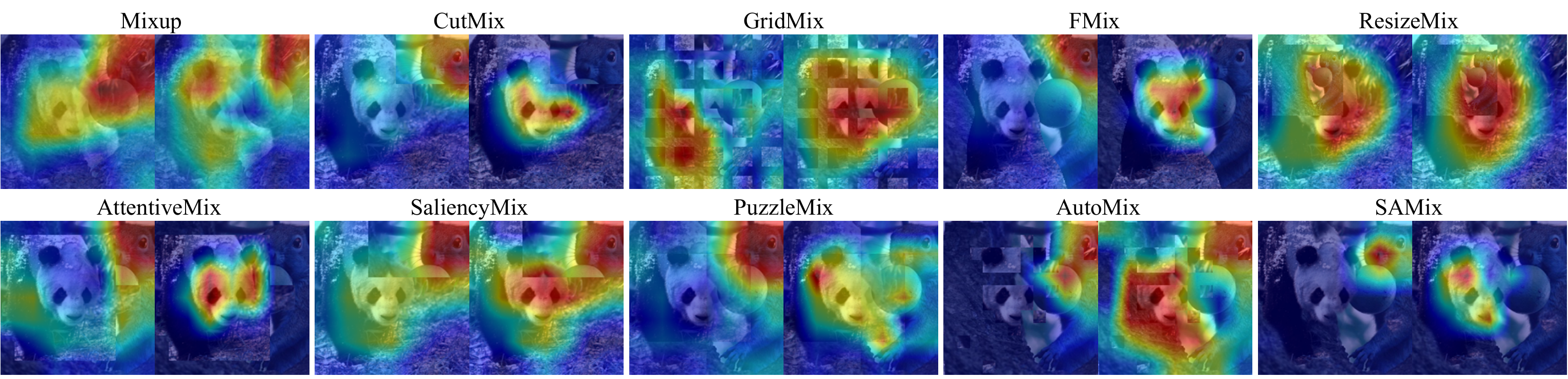}
    \vspace{-1.75em}
    \caption{Visualization of class activation mapping (CAM)~\citep{selvaraju2017gradcam} for top-1 and top-2 predicted classes of supported mixup methods with ResNet-50 on ImageNet-1K. Comparing the first and second rows, we observe that saliency-guided or dynamic mixup approaches (\textit{e.g.}, PuzzleMix and SAMix) localize the target regions better than the static methods (\textit{e.g.}, Mixup and ResizeMix).}
    \label{fig:vis_gradcam}
    \vspace{-1.0em}
\end{figure*}

\textbf{(G) Downstream Transferability \& CAM Visualization.}
To further evaluate the downstream performance and transferability of different mixup methods, we conduct transfer learning experiments on object detection~\citep{ren2015faster}, semantic segmentation~\citep{cvpr2019semanticFPN}, and weakly supervised object localization~\citep{cvpr2020MaxBoxAcc} with details in Appendix~\ref{app:transfer}. Notably, Table~\ref{tab:transfer_coco_cnn}, Table~\ref{tab:transfer_coco_vit}, and Table~\ref{tab:app_cub_wsol} suggest that \textit{dynamic} sampling mixing methods like AutoMix indeed exhibit competitive results, while recently proposed ViT-specific label mixing methods like TransMix perform even better, showcasing their superior transferability. The results also show the potential for improved online training mixup design. Moreover, it is commonly conjectured that vision models with better CAM localization could potentially be better transferred to fine-grained downstream prediction tasks.
As such, to gain intuitive insights, we also provide tools for class activation mapping (CAM) visualization with predicted classes in our codebase. As shown in Figure~\ref{fig:vis_gradcam} and Table~\ref{tab:app_cub_wsol}, \textit{dynamic} mixup like SAMix and AutoMix shows exceptional CAM localization, combined with their aforementioned accuracy, generalization ability, and robustness, may indicate their practical superiority compared to the \textit{static} ones in object detection and even borader downstream tasks.

% \vspace{-0.5em}
% \end{enumerate}

\section{Conclusion and Discussion}
\label{sec:conclusion}
% \paragraph{Contributions}
\textbf{Contributions.}
This paper presents OpenMixup, the \textit{first} comprehensive mixup augmentation benchmark and open-source codebase for visual representation learning, where 18 mixup algorithms are trained and thoroughly evaluated on 11 diverse vision datasets. 
The released codebase not only bolsters the entire benchmark but can facilitate broader under-explored mixup applications and downstream tasks.
Furthermore, observations and insights are obtained through different aspects of empirical analysis that are previously under-explored, such as GPU memory usage, loss landscapes, PL exponent alpha metrics, and more, contributing to a deeper and more systematic comprehension of mixup augmentation.
We anticipate that our OpenMixup benchmark and codebase can further contribute to fair and reproducible mixup research and we also encourage researchers and practitioners in the community to extend their valuable feedback to us and contribute to OpenMixup for building a more constructive mixup-based visual representation learning codebase together through GitHub.

% \vspace{-0.5em}
% \paragraph{Limitations and Future Works}
\textbf{Limitations and Future Works.}
The benchmarking scope of this work mainly focuses on visual classification, albeit we have supported a broader range of tasks in the proposed codebase and have conducted transfer learning experiments to object detection and semantic segmentation tasks to draw preliminary conclusions.
%While we posit that it is sufficient to characterize the mixup properties from classification, there is indeed limited coverage of downstream applications in this work. 
We are aware of this and have prepared it upfront for future work.
For example, our codebase can be easily extended to other computer vision tasks and datasets for further mixup benchmarking experiments and evaluations if necessary. Moreover, our observations and insights can also be of great value to the community for a more comprehensive understanding of mixup augmentation techniques.
We believe this work as the \textit{first} mixup benchmarking study is enough to serve as a kick-start, and we plan to extend our work in these directions in the future.

\section*{Acknowledgement}
This work was supported by National Key R\&D Program of China (No. 2022ZD0115100), National Natural Science Foundation of China Project (No. U21A20427), and Project (No. WU2022A009) from the Center of Synthetic Biology and Integrated Bioengineering of Westlake University. This work was done when Zedong Wang, Juanxi Tian, and Weiyang Jin interned at Westlake University. We sincerely thank Xin Jin for supporting evaluation implementations and all reviewers for polishing our manuscript. We also thank the AI Station of Westlake University for the support of GPUs.

%%%%%%%%% REFERENCES
{
\bibliography{ref}
\bibliographystyle{iclr2025_conference}
}

%%%%%%%%% APPENDIX
\clearpage
\renewcommand\thefigure{A\arabic{figure}}
\renewcommand\thetable{A\arabic{table}}
\setcounter{table}{0}
\setcounter{figure}{0}

\appendix

\onecolumn

\section*{Supplement Material}
In supplement material, we provide implementation details and full benchmark results of image classification, downstream tasks, and empirical analysis with mixup augmentations implemented in \texttt{OpenMixup} on various datasets.

% % % arXiv
% \section{Implementation Details}
% \label{app:implementation}
% \subsection{Setup OpenMixup}
% We simply introduce the installation and data preparation for \href{https://github.com/Westlake-AI/openmixup}{OpenMixup}, detailed in \href{https://openmixup.readthedocs.io/en/latest/install.html}{install.md}. Assuming the PyTorch environment has already been installed, users can easily reproduce the environment by executing the following commands:
% \vspace{-0.25em}
% \begin{lstlisting}[language=Python,
%   frame=single,
%   basicstyle=\small\ttfamily,
%   keywordstyle=\color{blue},
%   stringstyle=\color{red},
%   commentstyle=\color{black}]
% conda activate openmixup
% pip install openmim
% mim install mmcv-full
% git clone https://github.com/Westlake-AI/openmixup.git
% cd openmixup
% python setup.py develop  \# or "pip install -e ."
% \end{lstlisting}
% \vspace{-0.5em}
% Executing the instructions above, OpenMixup will be installed as the development mode, \textit{i.e.}, any modifications to the local source code take effect, and can be used as a python package.
% Then, users can download the datasets and \href{https://github.com/Westlake-AI/openmixup/releases/tag/dataset}{meta files}, and symlink them to the dataset root (\texttt{\$OpenMixup/data}).

% submit version
\section{Implementation Details}
\label{app:implementation}
\subsection{Setup OpenMixup}
As provided in the supplementary material or the 
% \href{https://openmixup.readthedocs.io/en/latest/}{online document}
\texttt{online document}, we simply introduce the installation and data preparation for OpenMixup,
detailed in ``docs/en/latest/install.md".
% detailed in \href{https://openmixup.readthedocs.io/en/latest/install.html}{docs/en/latest/install.md}.
Assuming the PyTorch environment has already been installed, users can easily reproduce the environment with the source code by executing the following commands:
\vspace{-0.25em}
\begin{lstlisting}[language=Python,
  frame=single,
  basicstyle=\small\ttfamily,
  keywordstyle=\color{blue},
  stringstyle=\color{red},
  commentstyle=\color{black}]
conda activate openmixup
pip install openmim
mim install mmcv-full
\# put the source code here
cd openmixup
python setup.py develop  \# or "pip install -e ."
\end{lstlisting}
\vspace{-0.5em}
Executing the instructions above, OpenMixup will be installed as the development mode, \textit{i.e.}, any modifications to the local source code take effect, and can be used as a python package. Then, users can download the datasets and the released meta files and symlink them to the dataset root (\texttt{\$OpenMixup/data}). The codebase is under \texttt{Apache 2.0} license.

\begin{figure*}[ht]
    \vspace{-0.5em}
    \centering
    \includegraphics[width=0.96\textwidth]{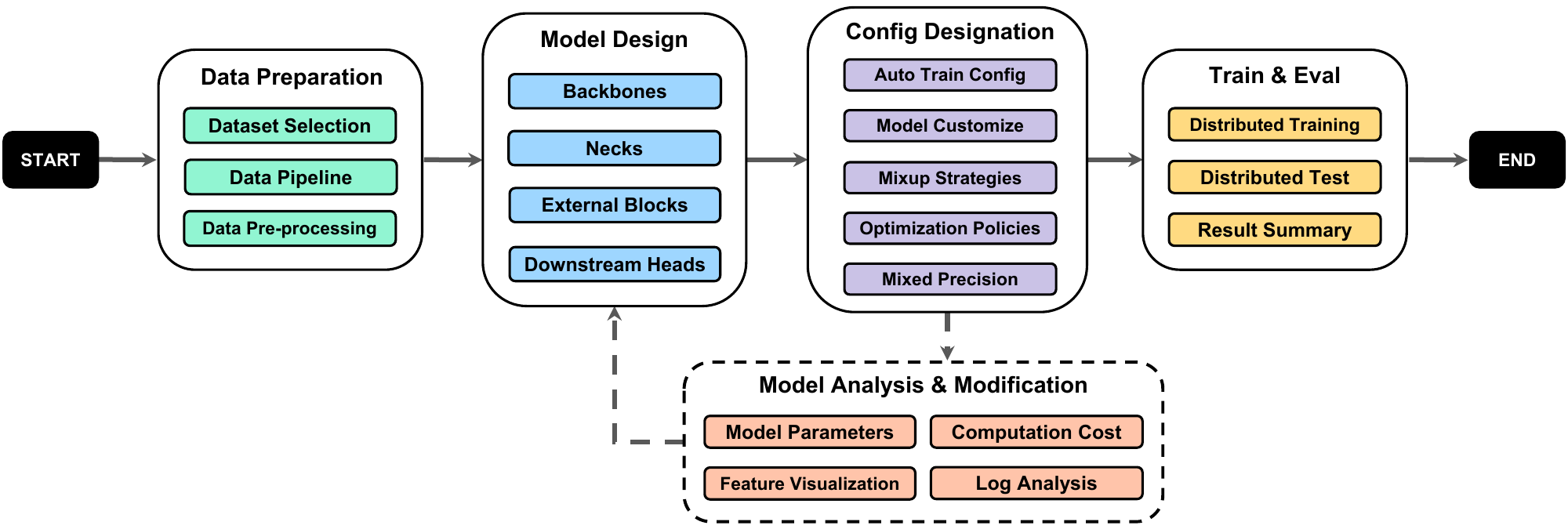}
    \vspace{-1.0em}
    \caption{Overview of the experimental pipeline in OpenMixup codebase.}
    \label{fig:pipline}
    \vspace{-0.75em}
\end{figure*}

\subsection{Training Settings of Image Classification}
\label{app:settings}
\paragraph{Large-scale Datasets.}
Table~\ref{tab:app_train_settings} illustrates three popular training settings on large-scaling datasets like ImageNet-1K in detail: (1) PyTorch-style~\citep{nips2019pytorch}. (2) DeiT~\citep{icml2021deit}. (3) RSB A2/A3~\citep{Wightman2021rsb}. Notice that the step learning rate decay strategy is replaced by Cosine Scheduler~\citep{loshchilov2016sgdr}, and \texttt{ColorJitter} as well as \texttt{PCA lighting} are removed in PyTorch-style setting for better performances. DeiT and RSB settings adopt advanced augmentation and regularization techniques for Transformers, while RSB A3 is a simplified setting for fast training on ImageNet-1K.
For a fare comparison, we search the optimal hyper-parameter $\alpha$ in $Beta(\alpha, \alpha)$ from $\{0.1, 0.2, 0.5, 1, 2, 4\}$ for compared methods while the rest of the hyper-parameters follow the original papers.

\begin{table}[ht]
\centering
    \vspace{-1.0em}
    \caption{Ingredients and hyper-parameters used for ImageNet-1K training settings.}
    % \vspace{-4pt}
\resizebox{0.60\linewidth}{!}{
\begin{tabular}{l|cccc}
	\toprule
	Procedure                  & PyTorch   & DeiT              & RSB A2            & RSB A3            \\ \hline
	Train Res                  & 224       & 224               & 224               & 160               \\
	Test Res                   & 224       & 224               & 224               & 224               \\
	Test  crop ratio           & 0.875     & 0.875             & 0.95              & 0.95              \\
	Epochs                     & 100/300   & 300               & 300               & 100               \\ \hline
	Batch size                 & 256       & 1024              & 2048              & 2048              \\
	Optimizer                  & SGD       & AdamW             & LAMB              & LAMB              \\
	LR                         & 0.1       & $1\times 10^{-3}$ & $5\times 10^{-3}$ & $8\times 10^{-3}$ \\
	LR decay                   & cosine    & cosine            & cosine            & cosine            \\
	Weight decay               & $10^{-4}$ & 0.05              & 0.02              & 0.02              \\
	Warmup epochs              & \xmarkg   & 5                 & 5                 & 5                 \\ \hline
	Label smoothing $\epsilon$ & \xmarkg   & 0.1               & \xmarkg           & \xmarkg           \\
	Dropout                    & \xmarkg   & \xmarkg           & \xmarkg           & \xmarkg           \\
	Stoch. Depth               & \xmarkg   & 0.1               & 0.05              & \xmarkg           \\
	Repeated Aug               & \xmarkg   & \cmark            & \cmark            & \xmarkg           \\
	Gradient Clip.             & \xmarkg   & 1.0               & \xmarkg           & \xmarkg           \\ \hline
	H. flip                    & \cmark    & \cmark            & \cmark            & \cmark            \\
	RRC                        & \cmark    & \cmark            & \cmark            & \cmark            \\
	Rand Augment               & \xmarkg   & 9/0.5             & 7/0.5             & 6/0.5             \\
	Auto Augment               & \xmarkg   & \xmarkg           & \xmarkg           & \xmarkg           \\
	Mixup alpha                & \xmarkg   & 0.8               & 0.1               & 0.1               \\
	Cutmix alpha               & \xmarkg   & 1.0               & 1.0               & 1.0               \\
	Erasing prob.              & \xmarkg   & 0.25              & \xmarkg           & \xmarkg           \\
	ColorJitter                & \xmarkg   & \xmarkg           & \xmarkg           & \xmarkg           \\
	EMA                        & \xmarkg   & \cmark            & \xmarkg           & \xmarkg           \\ \hline
	CE loss                    & \cmark    & \cmark            & \xmarkg           & \xmarkg           \\
	BCE loss                   & \xmarkg   & \xmarkg           & \cmark            & \cmark            \\ \hline
	Mixed precision            & \xmarkg   & \xmarkg           & \cmark            & \cmark            \\
    \bottomrule
    \end{tabular}
    }
\label{tab:app_train_settings}
\vspace{-1.0em}
\end{table}

\paragraph{Small-scale Datasets.}
We also provide two experimental settings on small-scale datasets:
(a) Following the common setups~\citep{He2016DeepRL, Yun2019CutMixRS} on small-scale datasets like CIFAR-10/100, we train 200/400/800/1200 epochs from stretch based on CIFAR version of ResNet variants~\citep{He2016DeepRL}, \textit{i.e.}, replacing the $7\times 7$ convolution and MaxPooling by a $3\times 3$ convolution. As for the data augmentation, we apply \texttt{RandomFlip} and \texttt{RandomCrop} with 4 pixels padding for 32$\times$32 resolutions. The testing image size is 32$\times$32 (no \texttt{CenterCrop}).
The basic training settings include: SGD optimizer with SGD weight decay of 0.0001, a momentum of 0.9, a batch size of 100, and a basic learning rate is 0.1 adjusted by Cosine Scheduler~\citep{loshchilov2016sgdr}. 
(b) We also provide modern training settings following DeiT~\citep{icml2021deit}, while using $224\times 224$ and $32\times 32$ resolutions for Transformer and CNN architectures. We only changed the batch size to 100 for CIFAR-100 and borrowed other settings the same as DeiT on ImageNet-1K.

% \paragraph{Tiny-ImageNet}
% This benchmark largely follows PuzzleMix~\citep{Kim2020PuzzleME} settings, which is similar to the CIFAR ones except for using \texttt{RandomFlip} and \texttt{RandomResizedCrop} for size 64$\times$64 as training augmentations. We adopt the basic learning rate of 0.2 and train 400 epochs with Cosine Scheduler, and the testing image size is 64$\times$64 (no \texttt{CenterCrop}).

% \paragraph{Downstream Tasks}
% We further provide benchmarks on three downstream classification scenarios: (a) Transfer learning on CUB-200 and FGVC-Aircraft. (b) Fine-grained classification on iNat2017 and iNat2018. (c) Scenic classification on Places205.
% For (a), we use transfer learning settings on fine-grained datasets, using PyTorch official pre-trained models as initialization and training 200 epochs by SGD optimizer with the initial learning rate of 0.001, the weight decay of 0.0005, the batch size of 16, the same data augmentation as ImageNet-1k settings.
% For (b) and (c), we follow Pytorch-style ImageNet-1k settings mentioned above, training 100 epochs from stretch.

\section{Mixup Image Classification Benchmarks}
\label{app:benchmark}

\subsection{Mixup Benchmarks on ImageNet-1k}
\label{app:in1k_benchmark}
\paragraph{PyTorch-style training settings}
The benchmark results are illustrated in Table~\ref{tab:cls_in_torch}. Notice that we adopt $\alpha=0.2$ for some cutting-based mixups (CutMix, SaliencyMix, FMix, ResizeMix) based on ResNet-18 since ResNet-18 might be under-fitted on ImageNet-1k.

% table: imagenet RSB & DeiTs
\begin{table*}[t!]
    \centering
    % \vspace{-0.5em}
    \caption{Top-1 accuracy (\%) of image classification based on ResNet variants on ImageNet-1K using PyTorch-style 100-epoch and 300-epoch training procedures.}
    % \vspace{-4pt}
\resizebox{0.92\linewidth}{!}{
    \begin{tabular}{l|c|ccccc|cccc}
    \toprule
                                      & Beta     & \multicolumn{5}{c|}{PyTorch 100 epochs} & \multicolumn{4}{c}{PyTorch 300 epochs} \\
                        Methods       & $\alpha$ & R-18   & R-34  & R-50  & R-101 & RX-101 & R-18     & R-34    & R-50    & R-101   \\ \hline
\rowcolor[HTML]{FDF0E2} Vanilla       & -        & 70.04  & 73.85 & 76.83 & 78.18 & 78.71  & 71.83    & 75.29   & 77.35   & 78.91   \\
\rowcolor[HTML]{FDF0E2} MixUp         & 0.2      & 69.98  & 73.97 & 77.12 & 78.97 & 79.98  & 71.72    & 75.73   & 78.44   & 80.60   \\
\rowcolor[HTML]{FDF0E2} CutMix        & 1        & 68.95  & 73.58 & 77.17 & 78.96 & 80.42  & 71.01    & 75.16   & 78.69   & 80.59   \\
\rowcolor[HTML]{FDF0E2} ManifoldMix   & 0.2      & 69.98  & 73.98 & 77.01 & 79.02 & 79.93  & 71.73    & 75.44   & 78.21   & 80.64   \\
\rowcolor[HTML]{FDF0E2} SaliencyMix   & 1        & 69.16  & 73.56 & 77.14 & 79.32 & 80.27  & 70.21    & 75.01   & 78.46   & 80.45   \\
\rowcolor[HTML]{FDF0E2} FMix          & 1        & 69.96  & 74.08 & 77.19 & 79.09 & 80.06  & 70.30    & 75.12   & 78.51   & 80.20   \\
\rowcolor[HTML]{FDF0E2} ResizeMix     & 1        & 69.50  & 73.88 & 77.42 & 79.27 & 80.55  & 71.32    & 75.64   & 78.91   & 80.52   \\
\rowcolor[HTML]{E7ECE4} PuzzleMix     & 1        & 70.12  & 74.26 & 77.54 & 79.43 & 80.53  & 71.64    & 75.84   & 78.86   & 80.67   \\
\rowcolor[HTML]{E7ECE4} AutoMix       & 2        & 70.50  & 74.52 & 77.91 & 79.87 & 80.89  & 72.05    & 76.10   & 79.25   & 80.98   \\
\rowcolor[HTML]{E7ECE4} AdAutoMix     & 1        & 70.86  & 74.82 & 78.04 & 79.91 & \bf{81.09}  & -        & -       & -       & -       \\
\rowcolor[HTML]{E7ECE4} SAMix         & 2        & \bf{70.83}  & \bf{74.95} & \bf{78.06} & \bf{80.05} & 80.98  & \bf{72.27}    & \bf{76.28}   & \bf{79.39}   & \bf{81.10}   \\
    \bottomrule
    \end{tabular}
    }
    \vspace{-1.0em}
    \label{tab:cls_in_torch}
\end{table*}
\begin{table*}[t]
    \vspace{-1.0em}
    \caption{Top-1 accuracy (\%) on ImageNet-1K based on popular Transformer-based architectures using DeiT-S training settings. Notice that $\dag$ denotes reproducing results with the official implementation, while other results are implemented with OpenMixup. TransMix, TokenMix, and SMMix are specially designed for Transformers.}
    \setlength{\tabcolsep}{1.5mm}
    % \vspace{-4pt}
\resizebox{0.975\linewidth}{!}{
    \begin{tabular}{l|c|cccccccc}
    \toprule
                        Methods          & $\alpha$ & DeiT-T & DeiT-S & DeiT-B & PVT-T & PVT-S & Swin-T & ConvNeXt-T & MogaNet-T \\ \hline
\rowcolor[HTML]{FDF0E2} Vanilla          & -        & 73.91  & 75.66  & 77.09  & 74.67 & 77.76 & 80.21  & 79.22      & 79.25     \\
\rowcolor[HTML]{FDF0E2} DeiT             & 0.8, 1   & 74.50  & 79.80  & 81.83  & 75.10 & 78.95 & 81.20  & 82.10      & 79.02     \\
\rowcolor[HTML]{FDF0E2} MixUp            & 0.2      & 74.69  & 77.72  & 78.98  & 75.24 & 78.69 & 81.01  & 80.88      & 79.29     \\
\rowcolor[HTML]{FDF0E2} CutMix           & 0.2      & 74.23  & 80.13  & 81.61  & 75.53 & 79.64 & 81.23  & 81.57      & 78.37     \\
\rowcolor[HTML]{FDF0E2} ManifoldMix      & 0.2      & -      & -      & -      & -     & -     & -      & 80.57      & 79.07     \\
\rowcolor[HTML]{FDF0E2} AttentiveMix+    & 2        & 74.07  & 80.32  & 82.42  & 74.98 & 79.84 & 81.29  & 81.14      & 77.53     \\
\rowcolor[HTML]{FDF0E2} SaliencyMix      & 0.2      & 74.17  & 79.88  & 80.72  & 75.71 & 79.69 & 81.37  & 81.33      & 78.74     \\
\rowcolor[HTML]{FDF0E2} FMix             & 0.2      & 74.41  & 77.37  &        & 75.28 & 78.72 & 79.60  & 81.04      & 79.05     \\
\rowcolor[HTML]{FDF0E2} ResizeMix        & 1        & 74.79  & 78.61  & 80.89  & 76.05 & 79.55 & 81.36  & 81.64      & 78.77     \\
\rowcolor[HTML]{E7ECE4} PuzzleMix        & 1        & 73.85  & 80.45  & 81.63  & 75.48 & 79.70 & 81.47  & 81.48      & 78.12     \\
\rowcolor[HTML]{E7ECE4} AutoMix          & 2        & 75.52  & 80.78  & 82.18  & 76.38 & 80.64 & 81.80  & 82.28      & 79.43     \\
\rowcolor[HTML]{E7ECE4} SAMix            & 2        & \bf{75.83} & \bf{80.94} & 82.85  & \bf{76.60} & 80.78 & \bf{81.87} & \bf{82.35} & \bf{79.62} \\
\rowcolor[HTML]{D1F5FF} TransMix         & 0.8, 1   & 74.56      & 80.68      & 82.51       & 75.50      & 80.50      & 81.80      & -          & -          \\
\rowcolor[HTML]{D1F5FF} TokenMix$^\dag$  & 0.8, 1   & 75.31      & 80.80      & \bf{82.90}  & 75.60      & -          & 81.60      & -          & -          \\
\rowcolor[HTML]{D1F5FF} SMMix            & 0.8, 1   & 75.56      & 81.10      & 82.90       & 75.60      & \bf{81.03} & 81.80      & -          & -          \\
    \bottomrule
    \end{tabular}
    }
    \label{tab:cls_in_deit}
    \vspace{-1.0em}
\end{table*}

\begin{table*}[t]
    \centering
    \caption{Top-1 accuracy (\%) on ImageNet-1K based on classical ConvNets using RSB A2/A3 training settings, including ResNet, EfficientNet, and MobileNet.V2.}
    % \vspace{-4pt}
    \setlength{\tabcolsep}{1.6mm}
\resizebox{0.825\linewidth}{!}{
    \begin{tabular}{l|c|cc|cc|cc}
    \toprule
                        Backbones     & $Beta$   & R-50       & R-50       & Eff-B0     & Eff-B0     & Mob.V2 $1\times$ & Mob.V2 $1\times$ \\
                        Settings      & $\alpha$ & A3         & A2         & A3         & A2         & A3               & A2               \\ \hline
\rowcolor[HTML]{FDF0E2} RSB           & 0.1, 1   & 78.08      & 79.80      & 74.02      & 77.26      & 69.86            & 72.87            \\
\rowcolor[HTML]{FDF0E2} MixUp         & 0.2      & 77.66      & 79.39      & 73.87      & 77.19      & 70.17            & 72.78            \\
\rowcolor[HTML]{FDF0E2} CutMix        & 0.2      & 77.62      & 79.38      & 73.46      & 77.24      & 69.62            & 72.23            \\
\rowcolor[HTML]{FDF0E2} ManifoldMix   & 0.2      & 77.78      & 79.47      & 73.83      & 77.22      & 70.05            & 72.34            \\
\rowcolor[HTML]{FDF0E2} AttentiveMix+ & 2        & 77.46      & 79.34      & 72.16      & 75.95      & 67.32            & 70.30            \\
\rowcolor[HTML]{FDF0E2} SaliencyMix   & 0.2      & 77.93      & 79.42      & 73.42      & 77.67      & 69.69            & 72.07            \\
\rowcolor[HTML]{FDF0E2} FMix          & 0.2      & 77.76      & 79.05      & 73.71      & 77.33      & 70.10            & 72.79            \\
\rowcolor[HTML]{FDF0E2} ResizeMix     & 1        & 77.85      & 79.94      & 73.67      & 77.27      & 69.94            & 72.50            \\
\rowcolor[HTML]{E7ECE4} PuzzleMix     & 1        & 78.02      & 79.78      & 74.10      & 77.35      & 70.04            & 72.85            \\
\rowcolor[HTML]{E7ECE4} AutoMix       & 2        & 78.44      & 80.28      & 74.61      & 77.58      & 71.16            & 73.19            \\
\rowcolor[HTML]{E7ECE4} SAMix         & 2        & \bf{78.64} & \bf{80.40} & \bf{75.28} & \bf{77.69} & \bf{71.24}       & \bf{73.42}       \\
    \bottomrule
    \end{tabular}
    }
    \label{tab:cls_in_rsb}
    \vspace{-1.0em}
\end{table*}

\paragraph{DeiT training setting}
Table~\ref{tab:cls_in_deit} shows the benchmark results following DeiT training setting. Experiment details refer to Sec.~\ref{app:settings}. Notice that the performances of transformer-based architectures are more difficult to reproduce than ResNet variants, and the mean of the best performance in 3 trials is reported as their original paper. 
\paragraph{RSB A2/A3 training settings}
The RSB A2/A3 benchmark results based on ResNet-50, EfficientNet-B0, and MobileNet.V2 are illustrated in Table~\ref{tab:cls_in_rsb}. 
Training 300/100 epochs with the BCE loss on ImageNet-1k, RSB A3 is a fast training setting, while RSB A2 can exploit the full representation ability of ConvNets. 
Notice that the RSB settings employ Mixup with $\alpha=0.1$ and CutMix with $\alpha=1.0$. We report the mean of top-1 accuracy in the last 5/10 training epochs for 100/300 epochs.

\subsection{Small-scale Classification Benchmarks}
\label{app:cifar_benchmark}
To facilitate fast research on mixup augmentations, we benchmark mixup image classification on CIFAR-10/100 and Tiny-ImageNet with two settings.

\paragraph{CIFAR-10}
As elucidated in Sec.~\ref{app:settings}, CIFAR-10 benchmarks based on CIFAR version ResNet variants follow CutMix settings, training 200/400/800/1200 epochs from stretch. As shown in Table~\ref{tab:cls_cifar10}, we report the median of top-1 accuracy in the last 10 training epochs.

% table: cifar-10
\begin{table*}[t!]
    \centering
    \vspace{-0.5em}
    \caption{Top-1 accuracy (\%) on CIFAR-10 training 200, 400, 800, 1200 epochs based on ResNet (R) and ResNeXt-32x4d (RX).}
    % \vspace{-4pt}
    \setlength{\tabcolsep}{1.3mm}
\resizebox{0.94\linewidth}{!}{
\begin{tabular}{l|c|cccc|c|cccc}
    \toprule
                        Backbones     & Beta     & R-18       & R-18       & R-18       & R-18       & Beta     & RX-50      & RX-50      & RX-50      & RX-50      \\
                        Epochs        & $\alpha$ & 200 ep     & 400 ep     & 800 ep     & 1200ep     & $\alpha$ & 200 ep     & 400 ep     & 800 ep     & 1200ep     \\ \hline
\rowcolor[HTML]{FDF0E2} Vanilla       & -        & 94.87      & 95.10      & 95.50      & 95.59      & -        & 95.92      & 95.81      & 96.23      & 96.26      \\
\rowcolor[HTML]{FDF0E2} MixUp         & 1        & 95.70      & 96.55      & 96.62      & 96.84      & 1        & 96.88      & 97.19      & 97.30      & 97.33      \\
\rowcolor[HTML]{FDF0E2} CutMix        & 0.2      & 96.11      & 96.13      & 96.68      & 96.56      & 0.2      & 96.78      & 96.54      & 96.60      & 96.35      \\
\rowcolor[HTML]{FDF0E2} ManifoldMix   & 2        & 96.04      & 96.57      & 96.71      & 97.02      & 2        & 96.97      & 97.39      & 97.33      & 97.36      \\
\rowcolor[HTML]{FDF0E2} SmoothMix     & 0.5      & 95.29      & 95.88      & 96.17      & 96.17      & 0.2      & 95.87      & 96.37      & 96.49      & 96.77      \\
\rowcolor[HTML]{FDF0E2} AttentiveMix+ & 2        & 96.21      & 96.45      & 96.63      & 96.49      & 2        & 96.84      & 96.91      & 96.87      & 96.62      \\
\rowcolor[HTML]{FDF0E2} SaliencyMix   & 0.2      & 96.05      & 96.42      & 96.20      & 96.18      & 0.2      & 96.65      & 96.89      & 96.70      & 96.60      \\
\rowcolor[HTML]{FDF0E2} FMix          & 0.2      & 96.17      & 96.53      & 96.18      & 96.01      & 0.2      & 96.72      & 96.76      & 96.76      & 96.10      \\
\rowcolor[HTML]{FDF0E2} GridMix       & 0.2      & 95.89      & 96.33      & 96.56      & 96.58      & 0.2      & 97.18      & 97.30      & 96.40      & 95.79      \\
\rowcolor[HTML]{FDF0E2} ResizeMix     & 1        & 96.16      & 96.91      & 96.76      & 97.04      & 1        & 97.02      & 97.38      & 97.21      & 97.36      \\
\rowcolor[HTML]{E7ECE4} PuzzleMix     & 1        & 96.42      & 96.87      & 97.10      & 97.13      & 1        & 97.05      & 97.24      & 97.37      & 97.34      \\
\rowcolor[HTML]{E7ECE4} AutoMix       & 2        & 96.59      & 97.08      & 97.34      & 97.30      & 2        & 97.19      & 97.42      & 97.65      & 97.51      \\
\rowcolor[HTML]{E7ECE4} SAMix         & 2        & \bf{96.67} & \bf{97.16} & \bf{97.50} & \bf{97.41} & 2        & \bf{97.23} & \bf{97.51} & \bf{97.93} & \bf{97.74} \\
    \bottomrule
    \end{tabular}
    }
    \label{tab:cls_cifar10}
    % \vspace{-0.5em}
\end{table*}

% table: cifar100
\begin{table*}[t!]
    \centering
    % \vspace{-2.0em}
    \caption{Top-1 accuracy (\%) on CIFAR-100 training 200, 400, 800, 1200 epochs based on ResNet (R), Wide-ResNet (WRN), ResNeXt-32x4d (RX). Notice that $\dag$ denotes reproducing results with the official implementation, while other results are implemented with OpenMixup.}
    % \vspace{-4pt}
    \setlength{\tabcolsep}{1.4mm}
\resizebox{0.99\linewidth}{!}{
\begin{tabular}{l|c|cccc|cccc|c}
    \toprule

                        Backbones       & Beta     & R-18       & R-18       & R-18       & R-18       & RX-50      & RX-50      & RX-50      & RX-50      & WRN-28-8   \\
                        Epochs          & $\alpha$ & 200 ep     & 400 ep     & 800 ep     & 1200ep     & 200 ep     & 400 ep     & 800 ep     & 1200ep     & 400ep      \\ \hline
\rowcolor[HTML]{FDF0E2} Vanilla         & -        & 76.42      & 77.73      & 78.04      & 78.55      & 79.37      & 80.24      & 81.09      & 81.32      & 81.63      \\
\rowcolor[HTML]{FDF0E2} MixUp           & 1        & 78.52      & 79.34      & 79.12      & 79.24      & 81.18      & 82.54      & 82.10      & 81.77      & 82.82      \\
\rowcolor[HTML]{FDF0E2} CutMix          & 0.2      & 79.45      & 79.58      & 78.17      & 78.29      & 81.52      & 78.52      & 78.32      & 77.17      & 84.45      \\
\rowcolor[HTML]{FDF0E2} ManifoldMix     & 2        & 79.18      & 80.18      & 80.35      & 80.21      & 81.59      & 82.56      & 82.88      & 83.28      & 83.24      \\
\rowcolor[HTML]{FDF0E2} SmoothMix       & 0.2      & 77.90      & 78.77      & 78.69      & 78.38      & 80.68      & 79.56      & 78.95      & 77.88      & 82.09      \\
\rowcolor[HTML]{FDF0E2} SaliencyMix     & 0.2      & 79.75      & 79.64      & 79.12      & 77.66      & 80.72      & 78.63      & 78.77      & 77.51      & 84.35      \\
\rowcolor[HTML]{FDF0E2} AttentiveMix+   & 2        & 79.62      & 80.14      & 78.91      & 78.41      & 81.69      & 81.53      & 80.54      & 79.60      & 84.34      \\
\rowcolor[HTML]{FDF0E2} FMix            & 0.2      & 78.91      & 79.91      & 79.69      & 79.50      & 79.87      & 78.99      & 79.02      & 78.24      & 84.21      \\
\rowcolor[HTML]{FDF0E2} GridMix         & 0.2      & 78.23      & 78.60      & 78.72      & 77.58      & 81.11      & 79.80      & 78.90      & 76.11      & 84.24      \\
\rowcolor[HTML]{FDF0E2} ResizeMix       & 1        & 79.56      & 79.19      & 80.01      & 79.23      & 79.56      & 79.78      & 80.35      & 79.73      & 84.87      \\
\rowcolor[HTML]{E7ECE4} PuzzleMix       & 1        & 79.96      & 80.82      & 81.13      & 81.10      & 81.69      & 82.84      & 82.85      & 82.93      & 85.02      \\
\rowcolor[HTML]{E7ECE4} Co-Mixup$^\dag$ & 2        & 80.01      & 80.87      & 81.17      & 81.18      & 81.73      & 82.88      & 82.91      & 82.97      & 85.05      \\
\rowcolor[HTML]{E7ECE4} AutoMix         & 2        & 80.12      & 81.78      & 82.04      & 81.95      & 82.84      & 83.32      & 83.64      & 83.80      & 85.18      \\
\rowcolor[HTML]{E7ECE4} SAMix           & 2        & 81.21      & \bf{81.97} & 82.30      & \bf{82.41} & 83.81      & \bf{84.27} & \bf{84.42} & \bf{84.31} & \bf{85.50} \\
\rowcolor[HTML]{E7ECE4} AdAutoMix       & 1        & \bf{81.55} & \bf{81.97} & \bf{82.32} & -          & \bf{84.40} & 84.05      & 84.42      & -          & 85.32      \\
    \bottomrule
    \end{tabular}
    }
    \label{tab:cls_cifar100}
    \vspace{-1.0em}
\end{table*}

\begin{table*}[t!]
    \centering
    \caption{Top-1 accuracy (\%), GPU memory (G), and total training time (h) of 600 epochs on CIFAR-100 training 200 and 600 epochs based on DeiT-S, Swin-T, and ConvNeXt-T with the DeiT training setting. Notice that all methods are trained on a single A100 GPU to collect training times and GPU memory.}
    % \vspace{-4pt}
    \setlength{\tabcolsep}{0.9mm}
\resizebox{1.0\linewidth}{!}{
\begin{tabular}{l|c|cccc|cccc|cccc}
    \toprule
                        Methods         & $\alpha$ & \multicolumn{4}{c|}{DeiT-Small}       & \multicolumn{4}{c|}{Swin-Tiny}        & \multicolumn{4}{c}{ConvNeXt-Tiny}     \\
                                        &          & 200 ep & 600 ep & Mem.     & Time     & 200 ep & 600 ep & Mem.     & Time     & 200 ep & 600 ep & Mem.     & Time     \\ \hline
\rowcolor[HTML]{FDF0E2} Vanilla         & -        & 65.81  & 68.50  & 8.1      & 27       & 78.41  & 81.29  & 11.4     & 36       & 78.70  & 80.65  & 4.2      & 10       \\
\rowcolor[HTML]{FDF0E2} Mixup           & 0.8      & 69.98  & 76.35  & 8.2      & 27       & 76.78  & 83.67  & 11.4     & 36       & 81.13  & 83.08  & 4.2      & 10       \\
\rowcolor[HTML]{FDF0E2} CutMix          & 2        & 74.12  & 79.54  & 8.2      & 27       & 80.64  & 83.38  & 11.4     & 36       & 82.46  & 83.20  & 4.2      & 10       \\
\rowcolor[HTML]{FDF0E2} DeiT            & 0.8, 1   & 75.92  & 79.38  & 8.2      & 27       & 81.25  & 84.41  & 11.4     & 36       & 83.09  & 84.12  & 4.2      & 10       \\
\rowcolor[HTML]{FDF0E2} ManifoldMix     & 2        & -      & -      & 8.2      & 27       & -      & -      & 11.4     & 36       & 82.06  & 83.94  & 4.2      & 10       \\
\rowcolor[HTML]{FDF0E2} SmoothMix       & 0.2      & 67.54  & 80.25  & 8.2      & 27       & 66.69  & 81.18  & 11.4     & 36       & 78.87  & 81.31  & 4.2      & 10       \\
\rowcolor[HTML]{FDF0E2} SaliencyMix     & 0.2      & 69.78  & 76.60  & 8.2      & 27       & 80.40  & 82.58  & 11.4     & 36       & 82.82  & 83.03  & 4.2      & 10       \\
\rowcolor[HTML]{FDF0E2} AttentiveMix+   & 2        & 75.98  & 80.33  & 8.3      & 35       & 81.13  & 83.69  & 11.5     & 43       & 82.59  & 83.04  & 4.3      & 14       \\
\rowcolor[HTML]{FDF0E2} FMix            & 1        & 70.41  & 74.31  & 8.2      & 27       & 80.72  & 82.82  & 11.4     & 36       & 81.79  & 82.29  & 4.2      & 10       \\
\rowcolor[HTML]{FDF0E2} GridMix         & 1        & 68.86  & 74.96  & 8.2      & 27       & 78.54  & 80.79  & 11.4     & 36       & 79.53  & 79.66  & 4.2      & 10       \\
\rowcolor[HTML]{FDF0E2} ResizeMix       & 1        & 68.45  & 71.95  & 8.2      & 27       & 80.16  & 82.36  & 11.4     & 36       & 82.53  & 82.91  & 4.2      & 10       \\
\rowcolor[HTML]{E7ECE4} PuzzleMix       & 2        & 73.60  & 81.01  & 8.3      & 35       & 80.33  & 84.74  & 11.5     & 45       & 82.29  & 84.17  & 4.3      & 53       \\
\rowcolor[HTML]{E7ECE4} AlignMix        & 1        & -      & -      & -        & -        & 78.91  & 83.34  & 12.6     & 39       & 80.88  & 83.03  & 4.2      & 13       \\
\rowcolor[HTML]{E7ECE4} AutoMix         & 2        & 76.24  & 80.91  & 18.2     & 59       & 82.67  & 84.05  & 29.2     & 75       & 83.30  & 84.79  & 10.2     & 56       \\
\rowcolor[HTML]{E7ECE4} SAMix           & 2        & \bf{77.94} & \bf{82.49} & 21.3  & 58  & \bf{82.70} & \bf{84.74} & 29.3 & 75   & \bf{83.56} & \bf{84.98} & 10.3     & 57       \\
\rowcolor[HTML]{D1F5FF} TransMix        & 0.8, 1   & 76.17  & 79.33  & 8.4      & 28       & 81.33  & 84.45  & 11.5     & 37       & -      & -      & -        & -        \\
\rowcolor[HTML]{D1F5FF} SMMix           & 0.8, 1   & 74.49  & 80.05  & 8.4      & 28       & 81.55  & -      & 11.5     & 37       & -      & -      & -        & -        \\
\rowcolor[HTML]{D1F5FF} Decoupled (DeiT) & 0.8, 1  & 76.75  & 79.78  & 8.2      & 27       & 81.10  & 84.59  & 11.4     & 36       & 83.44  & 84.49  & 4.2      & 10       \\
    \bottomrule
    \end{tabular}
    }
    \label{tab:cls_cifar100_vit}
    \vspace{-1.0em}
\end{table*}

\begin{table*}[t!]
    \centering
    % \vspace{-1.0em}
    \caption{More evaluation metric (robustness and calibration) on CIFAR-100 with 200-epoch training, reporting top-1 accuracy (\%)$\uparrow$ (clean data, corruption data, and FGSM attacks) and calibration ECE (\%)$\downarrow$.
    }
    % \vspace{-4pt}
    \setlength{\tabcolsep}{0.9mm}
\resizebox{0.825\linewidth}{!}{
\begin{tabular}{l|c|cccc|cccc}
    \toprule
Methods                                & $\alpha$ & \multicolumn{4}{c|}{DeiT-Small}                        & \multicolumn{4}{c}{Swin-Tiny}                          \\
                                       &          & Clean     & Corruption & FGSM       & ECE$\downarrow$ & Clean      & Corruption & FGSM       & ECE$\downarrow$ \\ \hline
\rowcolor[HTML]{FDF0E2} Vanilla        & -        & 65.81      & 49.31      & 20.58      & 9.48            & 78.41      & 58.20      & 12.87      & 11.67           \\
\rowcolor[HTML]{FDF0E2} Mixup          & 0.8      & 69.98      & 55.85      & 17.65      & 7.38            & 76.78      & 59.11      & 15.03      & 13.89           \\
\rowcolor[HTML]{FDF0E2} CutMix         & 2        & 74.12      & 55.08      & 12.53      & 6.18            & 80.64      & 57.73      & 18.38      & 10.95           \\
\rowcolor[HTML]{FDF0E2} DeiT           & 0.8, 1   & 75.92      & 57.36      & 18.55      & 5.38            & 81.25      & 62.21      & 15.66      & 15.68           \\
\rowcolor[HTML]{FDF0E2} SmoothMix      & 0.2      & 67.54      & 52.42      & 15.07      & 30.59           & 66.69      & 49.69      & 9.79       & 27.10           \\
\rowcolor[HTML]{FDF0E2} SaliencyMix    & 0.2      & 69.78      & 51.14      & 17.31      & 5.45            & 80.40      & 58.43      & 15.29      & 10.49           \\
\rowcolor[HTML]{FDF0E2} AttentiveMix+  & 2        & 75.98      & 57.57      & 13.90      & 9.89            & 81.13      & 58.07      & 15.43      & 9.60            \\
\rowcolor[HTML]{FDF0E2} FMix           & 1        & 70.41      & 51.94      & 12.20      & 4.14            & 80.72      & 58.44      & 13.97      & 9.19            \\
\rowcolor[HTML]{FDF0E2} GridMix        & 1        & 68.86      & 51.11      & 8.43       & 4.09            & 78.54      & 57.78      & 11.07      & 9.37            \\
\rowcolor[HTML]{FDF0E2} ResizeMix      & 1        & 68.45      & 50.87      & 20.03      & 7.64            & 80.16      & 57.37      & 13.64      & 7.68            \\
\rowcolor[HTML]{E7ECE4} PuzzleMix      & 2        & 73.60      & 57.67      & 17.44      & 9.45            & 80.33      & 60.67      & 12.96      & 16.23           \\
\rowcolor[HTML]{E7ECE4} AlignMix       & 1        & -          & -          & -          & -               & 78.91      & 61.61      & 17.20      & \bf{1.92}       \\
\rowcolor[HTML]{E7ECE4} AutoMix        & 2        & 76.24      & 60.08      & 27.35      & 4.69            & 82.67      & \bf{64.10} & 23.62      & 9.19            \\
\rowcolor[HTML]{E7ECE4} SAMix          & 2        & \bf{77.94} & \bf{61.91} & \bf{30.35} & \bf{4.01}       & \bf{82.70} & 62.19      & \bf{23.66} & 7.85            \\
\rowcolor[HTML]{D1F5FF} TransMix       & 0.8, 1   & 76.17      & 59.89      & 22.48      & 8.28            & 81.33      & 62.53      & 18.90      & 16.47           \\
\rowcolor[HTML]{D1F5FF} SMMix          & 0.8, 1   & 74.49      & 59.96      & 22.85      & 8.34            & 81.55      & 62.86      & 19.14      & 16.81           \\
\rowcolor[HTML]{D1F5FF} Decoupled (DeiT) & 0.8, 1 & 76.75      & 59.89      & 22.48      & 8.28            & 81.10      & 62.25      & 16.54      & 16.16           \\
    \bottomrule
    \end{tabular}
    }
    \label{tab:cls_cifar100_vit_robust}
    \vspace{-1.0em}
\end{table*}

\vspace{-0.25em}
\paragraph{CIFAR-100}
As for the classical setting (a), CIFAR-100 benchmarks train 200/400/800/1200 epochs from the stretch in Table~\ref{tab:cls_cifar100}, similar to CIFAR-10. Notice that we set weight decay to 0.0005 for cutting-based methods (CutMix, AttentiveMix+, SaliencyMix, FMix, ResizeMix) for better performances when using ResNeXt-50 (32x4d) as the backbone.
As shown in Table~\ref{tab:cls_cifar100_vit} using the modern setting (b), we train three modern architectures for 200/600 epochs from the stretch. We resize the raw images to $224\times 224$ resolutions for DeiT-S and Swin-T while modifying the stem network as the CIFAR version of ResNet for ConvNeXt-T with $32\times 32$ resolutions.
As shown in Table~\ref{tab:cls_cifar100_vit_robust}, we further provided more metrics to evaluate the robustness and reliability~\citep{nips2021intriguing, song2023perturb}: evaluating top-1 accuracy on the corrupted version of CIFAR-100~\citep{hendrycks2019benchmarking}, applying FGSM attack~\citep{iclr2015fgsm}), and visualizing the prediction calibration~\citep{Verma2019ManifoldMB}.

\begin{figure*}[t]
% \vspace{-0.5em}
\centering
\begin{minipage}{0.39\linewidth}
% \begin{table}[t]
\begin{table}[H]
    \centering
    \vspace{-0.5em}
    \caption{Top-1 accuracy (\%) on Tiny based on ResNet (R) and ResNeXt-32x4d (RX). Notice that $\dag$ denotes reproducing results with the official implementation, while other results are implemented with OpenMixup.}
    % \vspace{-4pt}
\resizebox{1.0\linewidth}{!}{
    \begin{tabular}{l|c|cc}
    \toprule
Backbones       & $\alpha$ & R-18      & RX-50      \\ \hline
\rowcolor[HTML]{FDF0E2} Vanilla         & -        & 61.68     & 65.04      \\
\rowcolor[HTML]{FDF0E2} MixUp           & 1        & 63.86     & 66.36      \\
\rowcolor[HTML]{FDF0E2} CutMix          & 1        & 65.53     & 66.47      \\
\rowcolor[HTML]{FDF0E2} ManifoldMix     & 0.2      & 64.15     & 67.30      \\
\rowcolor[HTML]{FDF0E2} SmoothMix       & 0.2      & 66.65     & 69.65      \\
\rowcolor[HTML]{FDF0E2} AttentiveMix+   & 2        & 64.85     & 67.42      \\
\rowcolor[HTML]{FDF0E2} SaliencyMix     & 1        & 64.60     & 66.55      \\
\rowcolor[HTML]{FDF0E2} FMix            & 1        & 63.47     & 65.08      \\
\rowcolor[HTML]{FDF0E2} GridMix         & 0.2      & 65.14     & 66.53      \\
\rowcolor[HTML]{FDF0E2} ResizeMix       & 1        & 63.74     & 65.87      \\
\rowcolor[HTML]{E7ECE4} PuzzleMix       & 1        & 65.81     & 67.83      \\
\rowcolor[HTML]{E7ECE4} Co-Mixup$^\dag$ & 2        & 65.92     & 68.02      \\
\rowcolor[HTML]{E7ECE4} AutoMix         & 2        & 67.33     & 70.72      \\
\rowcolor[HTML]{E7ECE4} SAMix           & 2        & 68.89     & 72.18      \\
\rowcolor[HTML]{E7ECE4} AdAutoMix       & 1        & \bf{69.19} & \bf{72.89} \\
    \bottomrule
    \end{tabular}
    }
    \label{tab:cls_tiny}
    % \vspace{-1.0em}
\end{table}

\end{minipage}
~\begin{minipage}{0.59\linewidth}
% \begin{figure}[ht]
% \begin{wrapfigure}{r}{0.5\linewidth}
    \centering
    \includegraphics[width=1.0\linewidth]{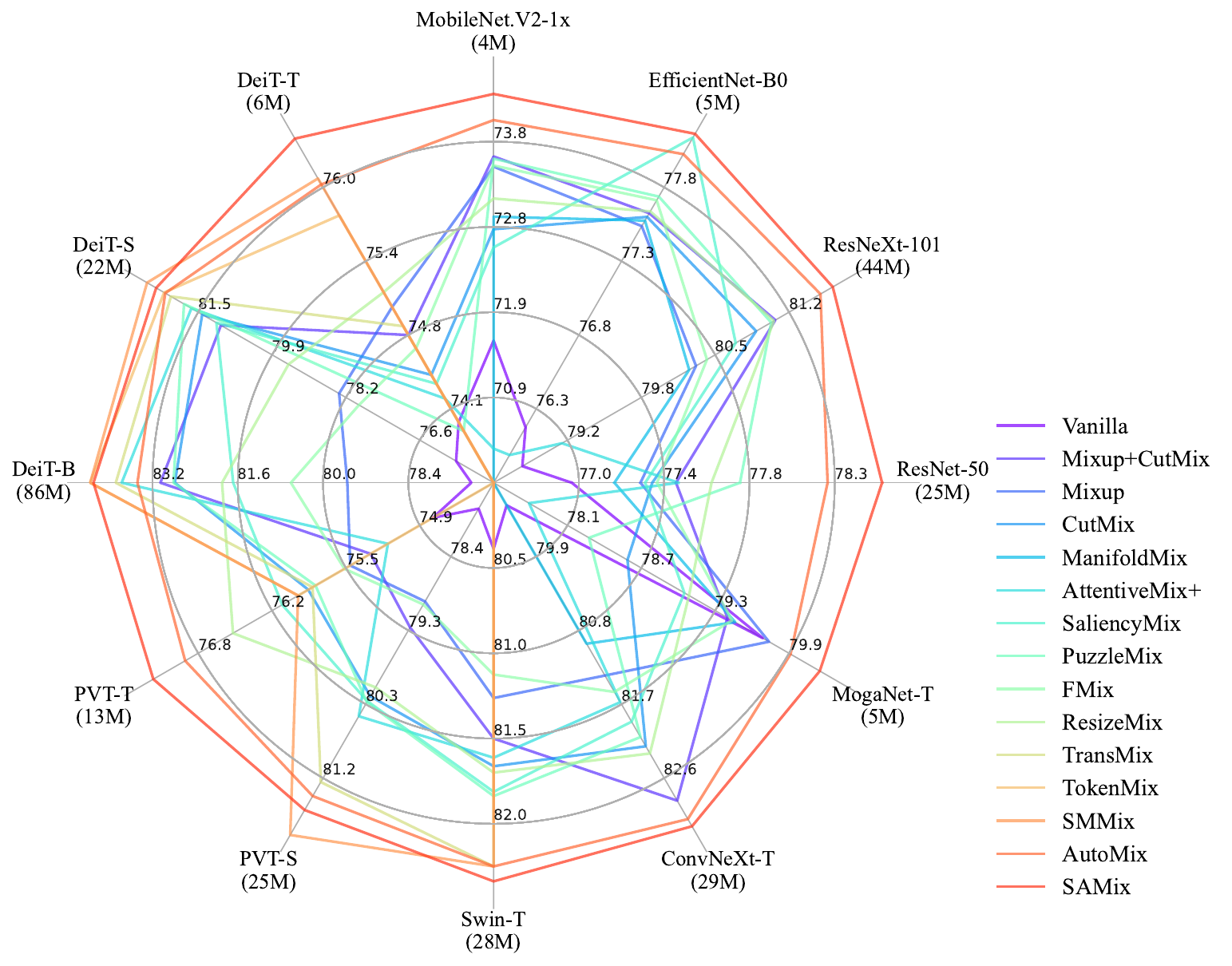}
    \vspace{-1.5em}
    \caption{Radar plots of the top-1 accuracy of all evaluated mixup augmentation methods based on a variety of popular vision backbones on ImageNet-1K.
    }
    \label{fig:in1k_loss_r50}
    \vspace{-1.0em}
% \end{wrapfigure}
% \end{figure}
\end{minipage}
\vspace{-1.0em}
\end{figure*}

% \vspace{-0.25em}
\paragraph{Tiny-ImageNet}
We largely follow the training setting of PuzzleMix~\citep{Kim2020PuzzleME} on Tiny-ImageNet, which adopts the basic augmentations of \texttt{RandomFlip} and \texttt{RandomResizedCrop} and optimize the models with a basic learning rate of 0.2 for 400 epochs with Cosine Scheduler.
As shown in Table~\ref{tab:cls_tiny}, all compared methods adopt ResNet-18 and ResNeXt-50 (32x4d) architectures training 400 epochs from the stretch on Tiny-ImageNet. 
% We search $\alpha$ in $Beta(\alpha, \alpha)$ for all compared methods.

% \subsection{Fine-grained Classification Benchmarks}
% \label{app:finegrained}
% \paragraph{Small-scale datasets.}
% We conduct small-scale fine-grained classification following transfer learning settings on CUB-200 and FGVC-Aircraft following transfer learning settings as presented in Sec.~\ref{app:settings}. We adopt official PyTorch pre-trained models on ImageNet-1k as initialization and report the median of Top-1 accuracy in the last 5/10 training epochs for 100/200 epochs. See the results in Table~\ref{tab:cls_fgvc_place}.

% \paragraph{Large-scale datasets.}
% For this benchmark, we adopt PyTorch-style ImageNet-1k settings as illustrated in Sec.~\ref{app:settings} with the total 100 training epochs from scratch on Places205 datasets based on ResNet variants as shown in Table~\ref{tab:cls_fgvc_place}. 

\subsection{Downstream Classification Benchmarks}
\label{app:downstream}
We further provide benchmarks on three downstream classification scenarios in 224$\times$224 resolutions with ResNet architectures, as shown in Table~\ref{tab:cls_fgvc_place}.

\vspace{-0.25em}
\paragraph{Benchmarks on Fine-grained Scenarios.}
As for fine-grained scenarios, each class usually has limited samples and is only distinguishable in some particular regions. We conduct (a) transfer learning on CUB-200 and FGVC-Aircraft and (b) fine-grained classification with training from scratch on iNat2017 and iNat2018.
For (a), we use transfer learning settings on fine-grained datasets, using PyTorch official pre-trained models as initialization and training 200 epochs by SGD optimizer with the initial learning rate of 0.001, the weight decay of 0.0005, the batch size of 16, the same data augmentation as ImageNet-1K settings. For (b) and (c), we follow Pytorch-style ImageNet-1K settings mentioned above, training 100 epochs from the stretch.

\vspace{-0.25em}
\paragraph{Benchmarks on Scenis Scenarios.}
As for scenic classification tasks, we study whether mixup augmentations help models distinguish the backgrounds, which are less important than the foreground objects in commonly used datasets. We employ the PyTorch-style training setting like ImageNet-1K on Places205~\citep{nips2014places205}, optimizing models for 100 epochs with SGD optimizer, a basic learning rate of 0.1 with 256 batch size.

\begin{table*}[t]
    \centering
    \vspace{-1.0em}
    \setlength{\tabcolsep}{1.1mm}
    \caption{Top-1 accuracy (\%) of mixup image classification with ResNet (R) and ResNeXt (RX) variants on fine-grained datasets (CUB-200, FGVC-Aircraft, iNat2017/2018) and Places205.}
    % \vspace{-4pt}
\resizebox{1.0\linewidth}{!}{
    \begin{tabular}{l|c|cc|cc|c|cc|cc|c|cc}
    \toprule
              & Beta     & \multicolumn{2}{c|}{CUB-200} & \multicolumn{2}{c|}{FGVC-Aircraft} & Beta     & \multicolumn{2}{c|}{iNat2017} & \multicolumn{2}{c|}{iNat2018} & Beta     & \multicolumn{2}{c}{Places205} \\
Method        & $\alpha$ & R-18          & RX-50        & R-18             & RX-50           & $\alpha$ & R-50          & RX-101        & R-50          & RX-101        & $\alpha$ & R-18          & R-50         \\ \hline
\rowcolor[HTML]{FDF0E2}Vanilla       & -        & 77.68         & 83.01        & 80.23            & 85.10           & -        & 60.23         & 63.70         & 62.53         & 66.94         & -        & 59.63      & 63.10        \\
\rowcolor[HTML]{FDF0E2}MixUp         & 0.2      & 78.39         & 84.58        & 79.52            & 85.18           & 0.2      & 61.22         & 66.27         & 62.69         & 67.56         & 0.2      & 59.33      & 63.01        \\
\rowcolor[HTML]{FDF0E2}CutMix        & 1        & 78.40         & 85.68        & 78.84            & 84.55           & 1        & 62.34         & 67.59         & 63.91         & 69.75         & 0.2      & 59.21      & 63.75        \\
\rowcolor[HTML]{FDF0E2}ManifoldMix   & 0.5      & 79.76         & 86.38        & 80.68            & 86.60           & 0.2      & 61.47         & 66.08         & 63.46         & 69.30         & 0.2      & 59.46      & 63.23        \\
\rowcolor[HTML]{FDF0E2}SaliencyMix   & 0.2      & 77.95         & 83.29        & 80.02            & 84.31           & 1        & 62.51         & 67.20         & 64.27         & 70.01         & 0.2      & 59.50      & 63.33        \\
\rowcolor[HTML]{FDF0E2}FMix          & 0.2      & 77.28         & 84.06        & 79.36            & 86.23           & 1        & 61.90         & 66.64         & 63.71         & 69.46         & 0.2      & 59.51      & 63.63        \\
\rowcolor[HTML]{FDF0E2}ResizeMix     & 1        & 78.50         & 84.77        & 78.10            & 84.0            & 1        & 62.29         & 66.82         & 64.12         & 69.30         & 1        & 59.66      & 63.88        \\
\rowcolor[HTML]{E7ECE4}PuzzleMix     & 1        & 78.63         & 84.51        & 80.76            & 86.23           & 1        & 62.66         & 67.72         & 64.36         & 70.12         & 1        & 59.62      & 63.91        \\
\rowcolor[HTML]{E7ECE4}AutoMix       & 2        & 79.87         & 86.56        & 81.37            & 86.72           & 2        & 63.08         & 68.03         & 64.73         & 70.49         & 2        & 59.74      & 64.06        \\
\rowcolor[HTML]{E7ECE4}SAMix         & 2        & \bf{81.11}    & \bf{86.83}   & \bf{82.15}       & \bf{86.80}      & 2        & \bf{63.32}    & \bf{68.26}    & \bf{64.84}    & \bf{70.54}    & 2        & \bf{59.86} & \bf{64.27}    \\
    \bottomrule
    \end{tabular}
    }
    \label{tab:cls_fgvc_place}
    \vspace{-0.5em}
\end{table*}

\begin{figure*}[ht]  % Top
    \vspace{-0.5em}
\centering
    \subfigtopskip=-0.5pt
    \subfigbottomskip=-0.5pt
    \subfigcapskip=-4pt
    % \hspace{-0.5em}
    % \subfigure[Deit-S (200 epochs) on CIFAR-100]{\label{fig:ep_acc_cifar_deit_s}\includegraphics[width=0.48\linewidth,trim= 0 0 0 0,clip]{images/fig_acc_plot_cifar100_deit_s_200ep.pdf}}
    % \subfigure[Swin-T (200 epochs) on CIFAR-100]{\label{fig:ep_acc_cifar_swin_t}\includegraphics[width=0.48\linewidth,trim= 0 0 0 0,clip]{images/fig_acc_plot_cifar100_swin_t_200ep.pdf}}
    % 
    \hspace{-0.5em}
    \subfigure[ResNet-50 on ImageNet-1K]{\label{fig:ep_acc_in1k_r50_100ep}\includegraphics[width=0.48\linewidth,trim= 0 0 0 0,clip]{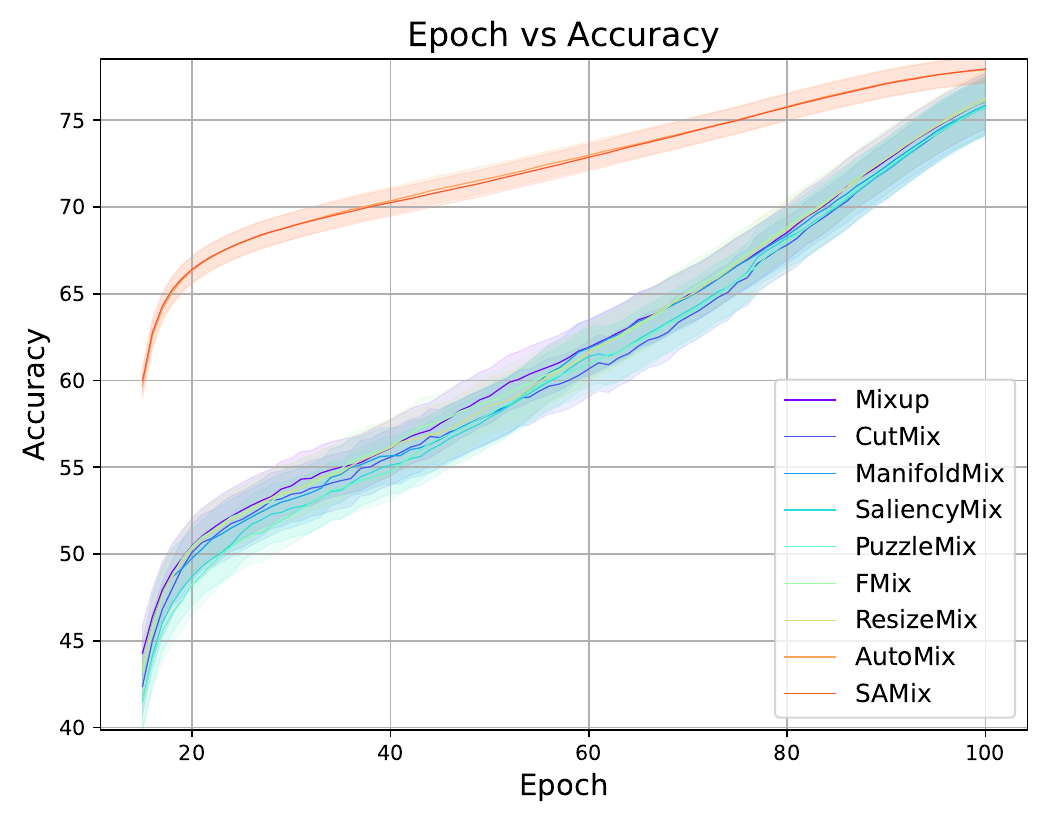}}
    \subfigure[Swin-T on ImageNet-1K]{\label{fig:ep_acc_in1k_swin_t}\includegraphics[width=0.48\linewidth,trim= 0 0 0 0,clip]{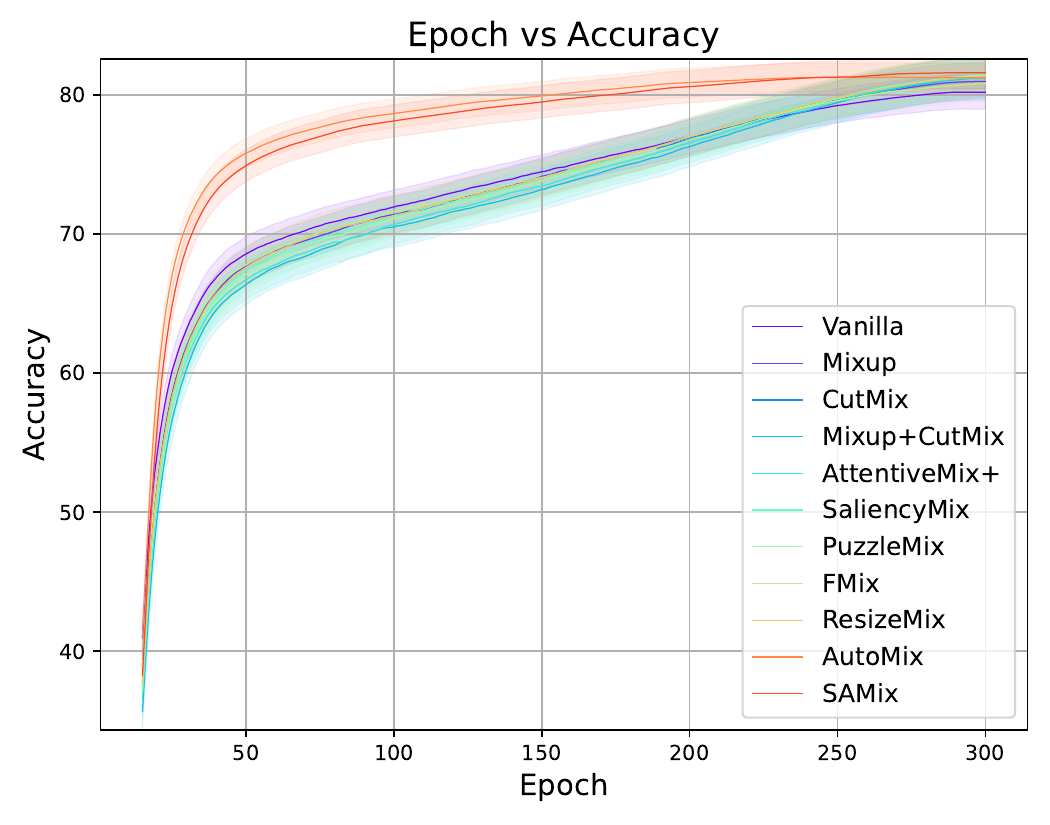}}
    \hspace{-0.5em}
    \subfigure[ResNet-50 on iNatualist2017]{\label{fig:ep_acc_inat2017_r50}\includegraphics[width=0.48\linewidth,trim= 0 0 0 0,clip]{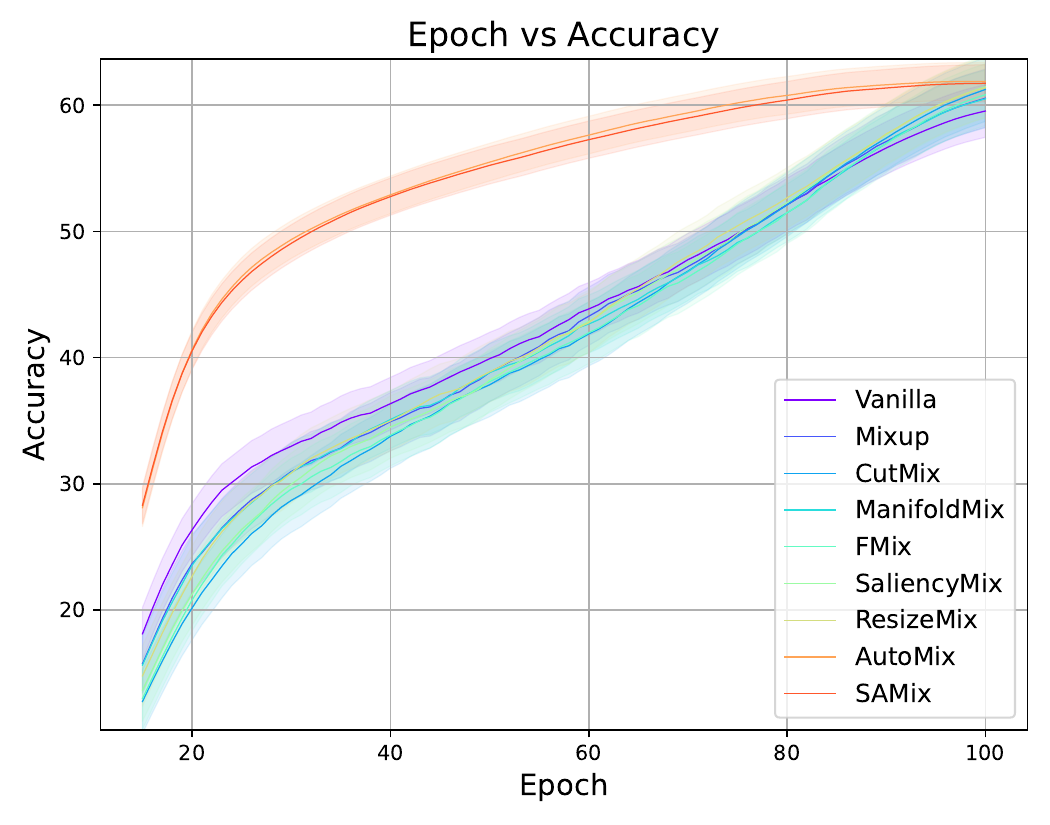}}
    \subfigure[ResNet-50 on Place205]{\label{fig:ep_acc_place205}\includegraphics[width=0.48\linewidth,trim= 0 0 0 0,clip]{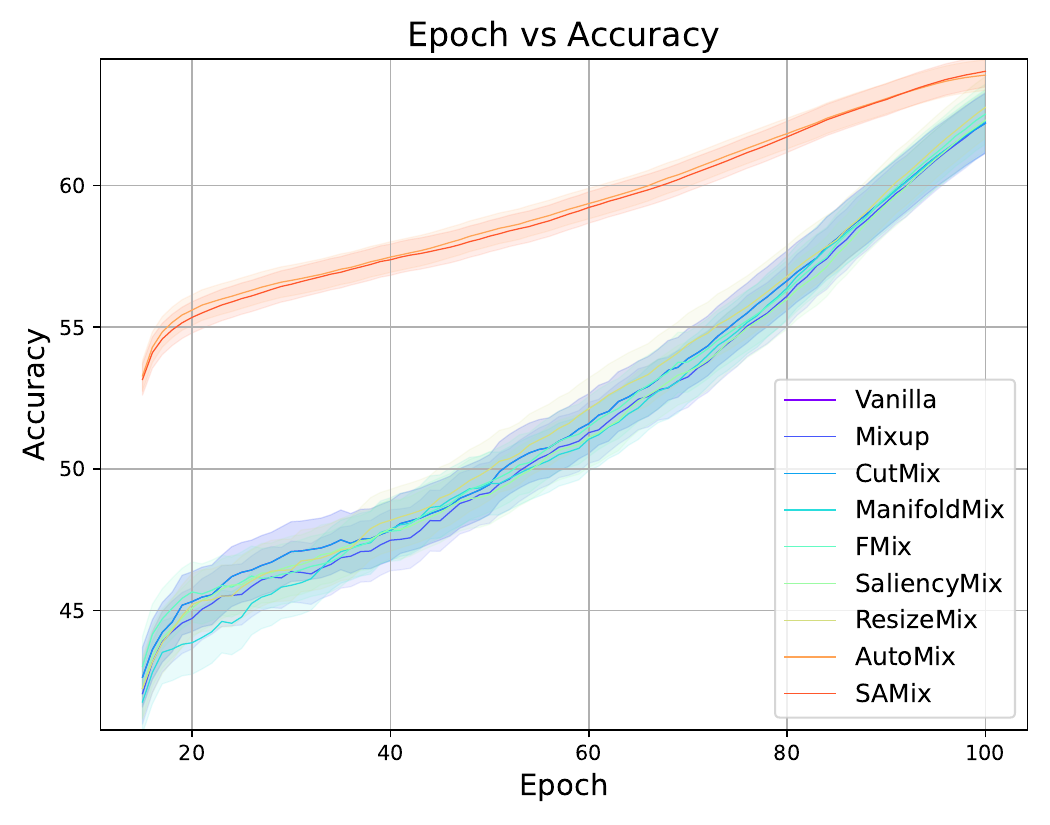}}
\vspace{-0.5em}
    \caption{Training epoch \textit{vs.} top-1 accuracy plots of various mixup methods on (a)(b) ImageNet-1K, (c) iNatualist2017, and (d) Place205 to further study training stability and convergence speed.
    }
    \label{fig:app_ep_vs_acc}
    \vspace{-1.5em}
\end{figure*}

\paragraph{Visualization of Training Stabiltities.}
To further analyze training stability and convergence speed, we provided more visualization of the training epoch vs. top-1 validation accuracy of various Mixup augmentations on different datasets to support the conclusion of training convergence, as shown in Figure~\ref{fig:app_ep_vs_acc}. These visualization results could be easily obtained by our analysis tools under {\color{purple}\texttt{tools/analysis\_tools}}.

\begin{figure*}[ht]  % Top
    % \vspace{-0.5em}
\centering
    \subfigtopskip=0.0pt
    \subfigbottomskip=2.5pt
    \subfigcapskip=-4pt
    % \hspace{-0.5em}
    \subfigure[R-50 on ImageNet-1K]{\label{fig:alpha_in1k_r50}\includegraphics[width=0.45\linewidth,trim= 0 0 0 0,clip]{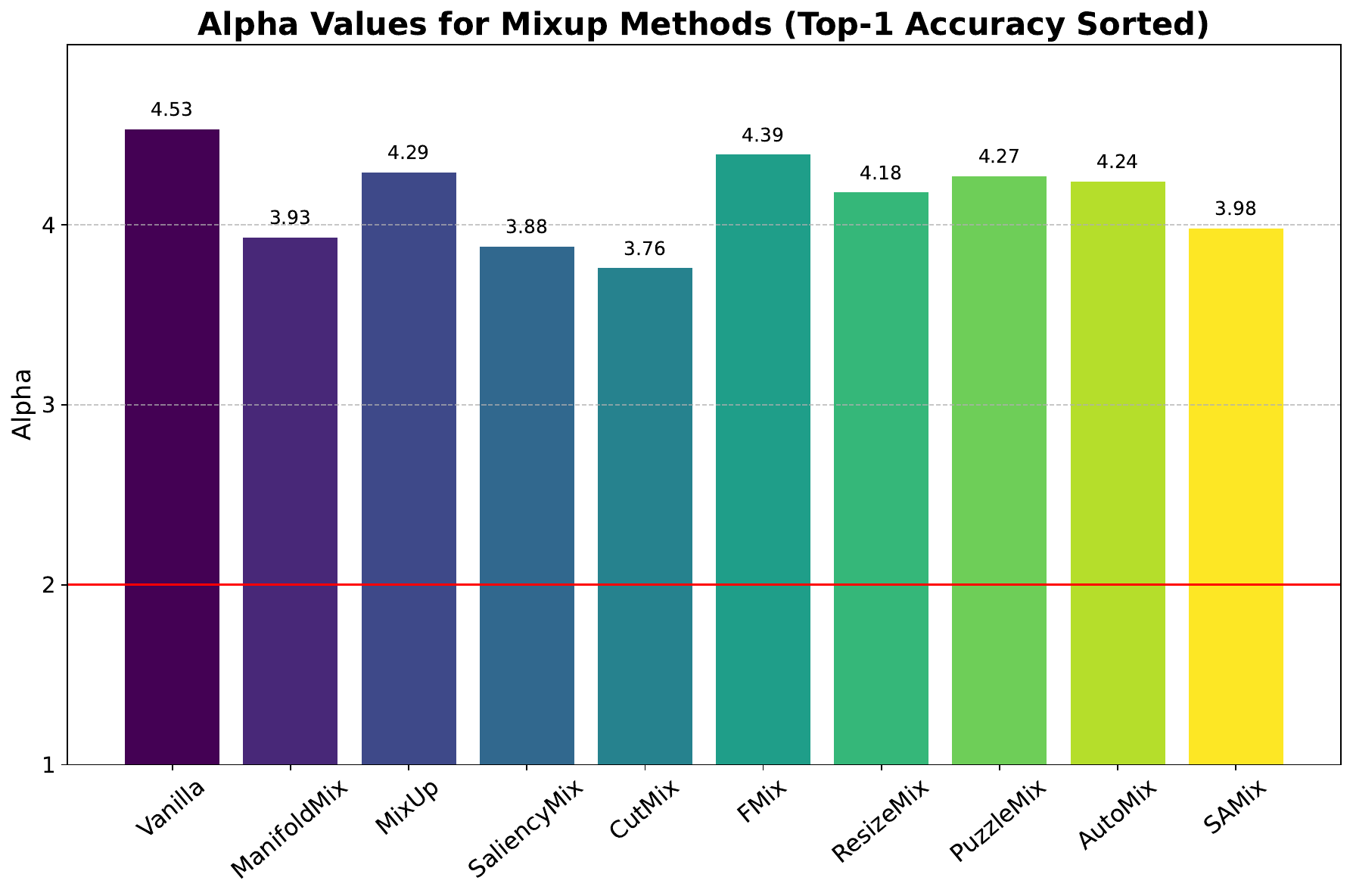}}
    \subfigure[Swin-T on ImageNet-1K]{\label{fig:alpha_in1k_swin}\includegraphics[width=0.45\linewidth,trim= 0 0 0 0,clip]{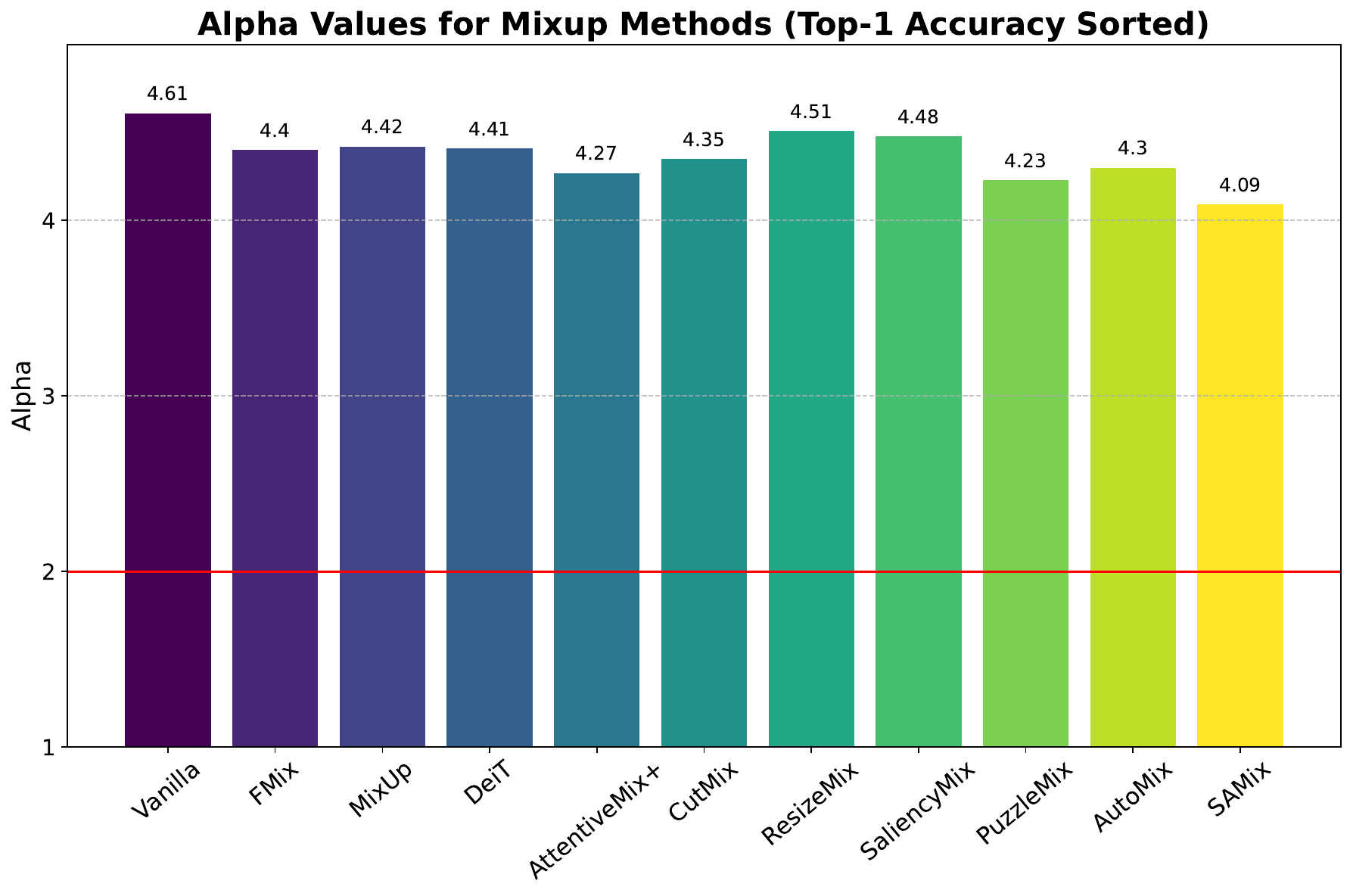}}
\newline
\centering
    \subfigure[R-50 on iNatural2017]{\hspace{-1.0em}\label{fig:alpha_inat2017_r50}\includegraphics[width=0.45\linewidth,trim= 0 0 0 0,clip]{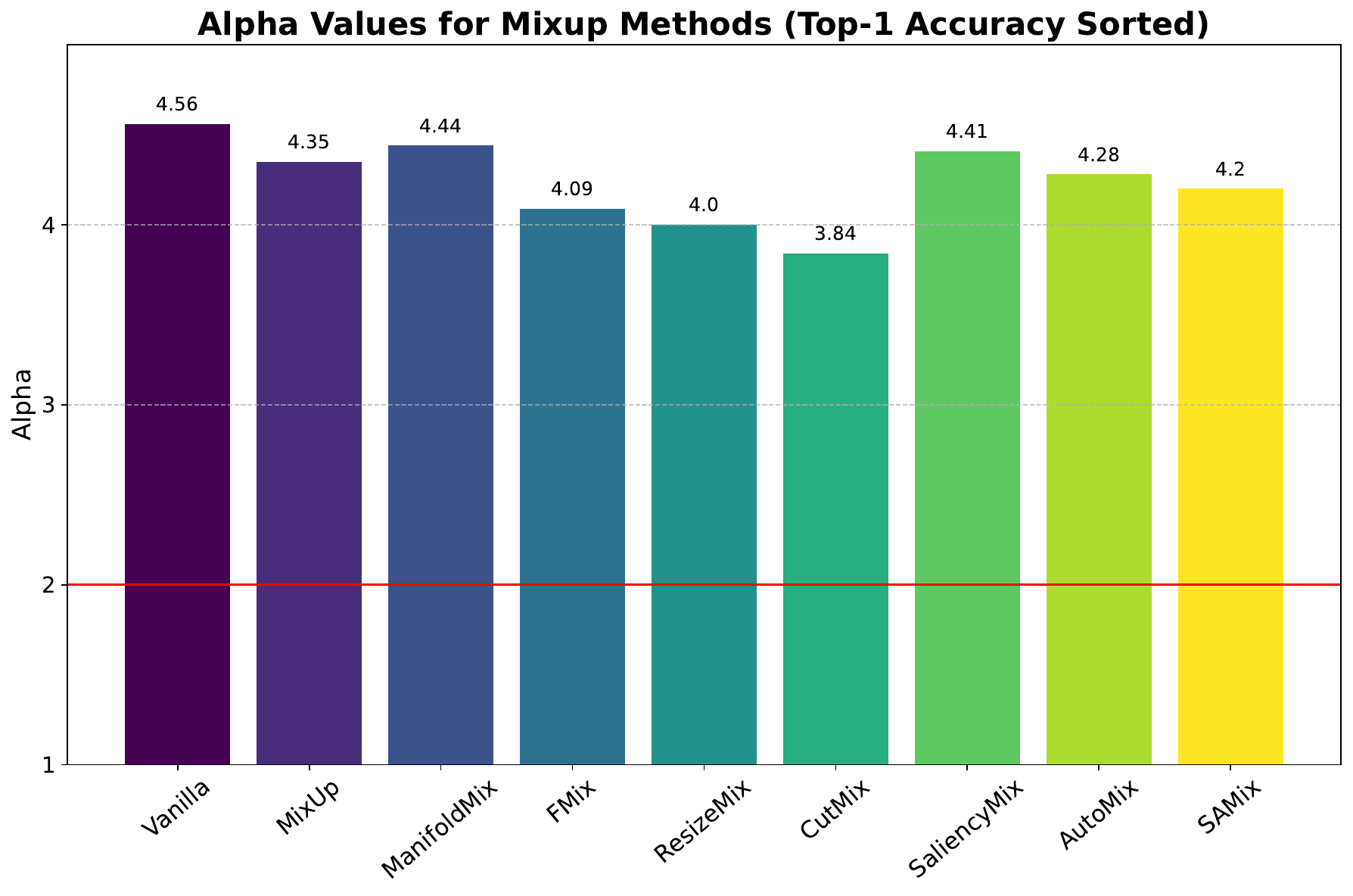}}
    \subfigure[R-50 on Place205]{\label{fig:alpha_place205_r50}\includegraphics[width=0.45\linewidth,trim= 0 0 0 0,clip]{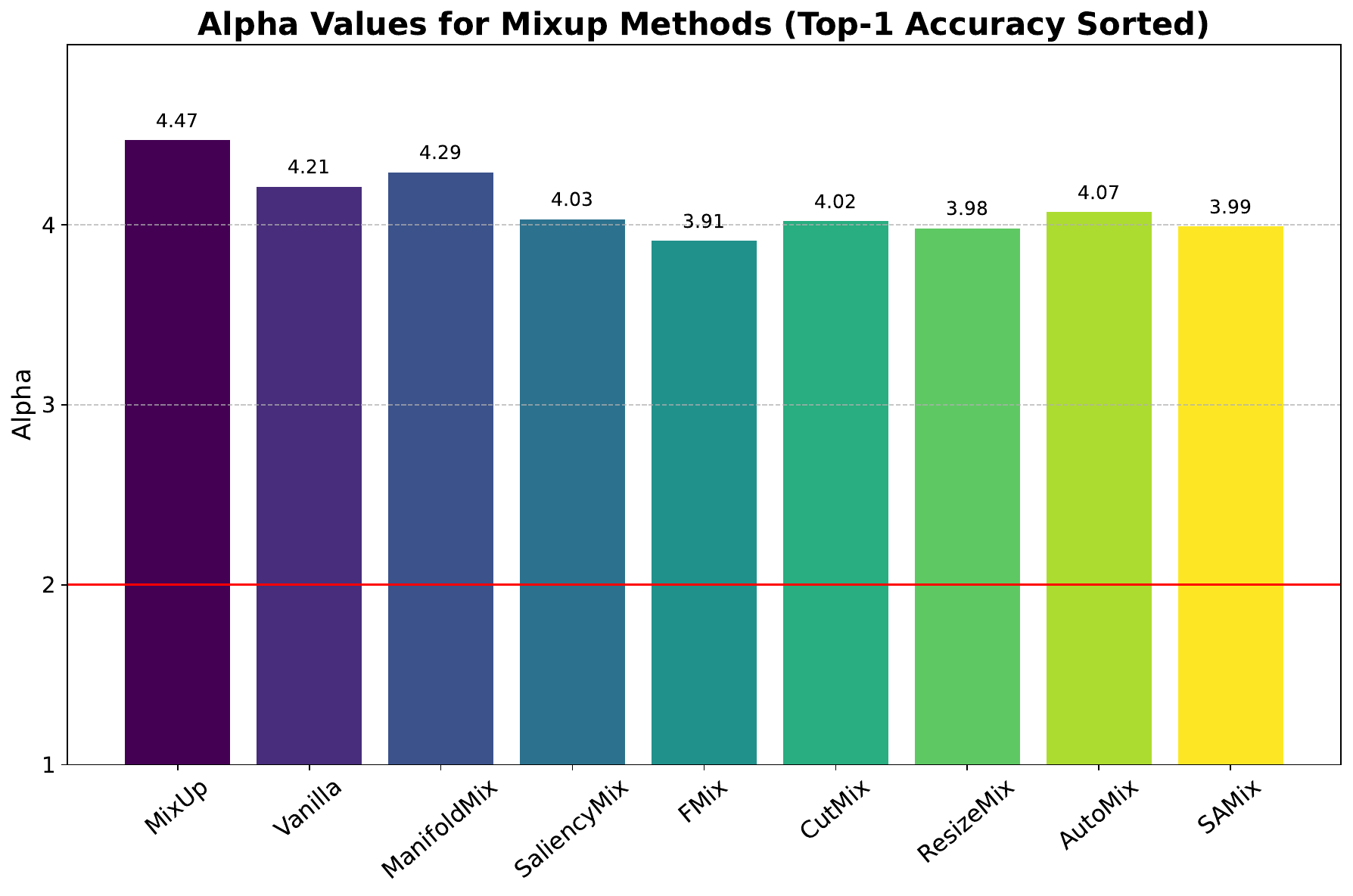}}
\vspace{-0.75em}
    \caption{Explaination of learned ResNet-50 or Swin-T by various mixup methods with alpha metrics computed by \texttt{WeightWather} on (a)(b) ImageNet-1K, and (c) iNaturalist2017, and (d) Place205. In each figure, the bars are sorted with the top-1 accuracy from left to right. Empirically, the alpha metric indicates the degree of how well models fit the task, where alpha less than 2 or greater than 6 indicates the risk of overfitting and underfitting. (a)(b) On ImageNet-1K, favorable mixup methods (\textit{e.g.,} \textit{dynamic} ones like AutoMix variants) prevent ResNet-50 (already had inductive bias) from overfitting while helping Swin-T learning better representations. (c) Since iNaturalist2017 is a smaller dataset with more difficult classes than ImageNet-1K, \textit{dynamic} mixup methods tend to prevent overfitting to get better fine-grained classification performances. (d) Place205 with difficult scenic images, is two times larger than ImageNet-1K with iconic images. Therefore, it is likely to require mixup augmentations to encourage better fitting to scenic classification.
    }
    \label{fig:app_alpha_norm}
    \vspace{-0.5em}
\end{figure*}

% \newpage

\subsection{Transfer Learning}
\label{app:transfer}
\paragraph{Object Detection.}
We conduct transfer learning experiments with pre-trained ResNet-50~\citep{He2016DeepRL} and PVT-S~\citep{iccv2021PVT} using mixup augmentations to object detection on COCO-2017~\citep{eccv2014MSCOCO} dataset, which evaluate the generalization abilities of different mixup approaches. We first fine-tune Faster RCNN~\citep{ren2015faster} with ResNet-50-C4 using Detectron2~\citep{wu2019detectron2} in Table~\ref{tab:transfer_coco_cnn}, which is trained by SGD optimizer and multi-step scheduler for 24 epochs ($2\times$). The \textit{dynamic} mixup methods (\textit{e.g., } AutoMix) usually achieve both competitive performances in classification and object detection tasks.
Then, we fine-tune Mask R-CNN~\citep{2017iccvmaskrcnn} by AdamW optimizer for 24 epochs using MMDetection~\citep{2019mmdetection} in Table~\ref{tab:transfer_coco_vit}. We have integrated Detectron2 and MMDetection into OpenMixup, and the users can perform the transferring experiments with pre-trained models and config files. Compared to \textit{dynamic} sample mixing methods, recently-proposed label mixing policies (\textit{e.g.,} TokenMix and SMMix) yield better performances with less extra training overheads.

\paragraph{Semantic Segmentation.}
We also perform transfer learning to semantic segmentation on ADE20K~\citep{Zhou2018ADE20k} with Semantic FPN~\citep{cvpr2019semanticFPN} to evaluate the generalization abilities to fine-grained prediction tasks. Following PVT~\citep{iccv2021PVT}, we fine-tuned Semantic FPN for 80K interactions by AdamW~\citep{iclr2019AdamW} optimizer with the learning rate of $2\times 10^{-4}$ and a batch size of 16 on $512^2$ resolutions using MMSegmentation~\citep{mmseg2020}. Table~\ref{tab:transfer_coco_vit} shows the results of transfer experiments based on PVT-S.

\begin{figure*}[t]
    \vspace{-2.0em}
\begin{minipage}{0.40\linewidth}
\centering
    % 02.28
\begin{table}[H]
\centering
    \setlength{\tabcolsep}{1.5mm}
    \caption{Trasfer learning of object detection with ImageNet-1k pre-trained ResNet-50 backbone on COCO dataset.}
    % \vspace{-0.5em}
\resizebox{1.0\linewidth}{!}{
\begin{tabular}{l|c|ccc}
    \toprule
                                  & IN-1K     & \multicolumn{3}{c}{COCO}                    \\
                        Method    & Acc       & mAP       & AP$^{bb}_{50}$ & AP$^{bb}_{75}$ \\ \hline
\rowcolor[HTML]{FDF0E2} Vanilla   & 76.8      & 38.1      & 59.1           & 41.8           \\
\rowcolor[HTML]{FDF0E2} Mixup     & 77.1      & 37.9      & 59.0           & 41.7           \\
\rowcolor[HTML]{FDF0E2} CutMix    & 77.2      & 38.2      & 59.3           & 42.0           \\
\rowcolor[HTML]{FDF0E2} ResizeMix & 77.4      & 38.4      & 59.4           & 42.1           \\
\rowcolor[HTML]{E7ECE4} PuzzleMix & 77.5      & 38.3      & 59.3           & 42.1           \\
\rowcolor[HTML]{E7ECE4} AutoMix   & 77.9      & 38.6      & 59.5           & \bf{42.2}      \\
\rowcolor[HTML]{E7ECE4} SAMix     & \bf{78.1} & \bf{38.7} & \bf{59.6}      & \bf{42.2}      \\
    \bottomrule
    \end{tabular}
    }
\label{tab:transfer_coco_cnn}
\end{table}

\end{minipage}
~\begin{minipage}{0.59\linewidth}
\centering
    % 02.28
\begin{table}[H]
\centering
    \setlength{\tabcolsep}{1.5mm}
    \caption{Trasfer learning of object detection with Mask R-CNN and semantic segmentation with Semantic FPN with pre-trained PVT-S on COCO and ADE20K, respectively.}
    % \vspace{-0.5em}
\resizebox{1.0\linewidth}{!}{
\begin{tabular}{l|c|ccc|c}
    \toprule
                                     & IN-1K     & \multicolumn{3}{c|}{COCO}                   & ADE20K    \\
Method                               & Acc       & mAP       & AP$^{bb}_{50}$ & AP$^{bb}_{75}$ & mIoU      \\ \hline
\rowcolor[HTML]{FDF0E2} MixUp+CutMix & 79.8      & 40.4      & 62.9           & 43.8           & 41.9      \\
\rowcolor[HTML]{E7ECE4} AutoMix      & 80.7      & 40.9      & 63.9           & 44.1           & 42.5      \\
\rowcolor[HTML]{D1F5FF} TransMix     & 80.5      & 40.9      & 63.8           & 44.0           & 42.6      \\
\rowcolor[HTML]{D1F5FF} TokenMix     & 80.6      & \bf{41.0} & \bf{64.0}      & 44.3           & \bf{42.7} \\
\rowcolor[HTML]{D1F5FF} TokenMixup   & 80.5      & 40.7      & 63.6           & 43.9           & 42.5      \\
\rowcolor[HTML]{D1F5FF} SMMix        & \bf{81.0} & \bf{41.0} & 63.9           & \bf{44.4}      & \bf{43.0} \\
    \bottomrule
    \end{tabular}
    }
\label{tab:transfer_coco_vit}
\end{table}

\end{minipage}
    \vspace{-0.5em}
\end{figure*}

\paragraph{Weakly Supervised Object Localization.}
To study the localization ability of trained models more precisely, we follow CutMix~\citep{Yun2019CutMixRS} to evaluate the weakly supervised object localization (WSOL) task on CUB-200~\citep{Wah2011CUB2011}. The model localizes objects of interest based on the activation maps of CAM~\citep{selvaraju2017gradcam} without bounding box supervision and calculates the maximal box accuracy with a threshold $\delta \in \{0.3, 0.5, 0.7\}$ as MaxBoxAccV2~\citep{cvpr2020MaxBoxAcc}. We provided the benchmarked results on CUB-200 in Table~\ref{tab:app_cub_wsol}, where we found similar conclusions as the visualization of Grad-CAM in Sec.~\ref{sec:observation}.

\begin{table*}[ht]
    \centering
    \vspace{-1.0em}
    \caption{
    MaxBoxAcc (\%)$\uparrow$ for the Weakly Supervised Object Localization (WSOL) task on CUB-200 based on ResNet architectures.
    Following CutMix~\citep{Yun2019CutMixRS}, the model localizes objects of interest based on the activation maps of CAM~\citep{selvaraju2017gradcam} without bounding box supervision and calculates the maximal box accuracy with a threshold $\delta \in \{0.3, 0.5, 0.7\}$ as MaxBoxAccV2~\citep{cvpr2020MaxBoxAcc}.
    }
    % \vspace{-4pt}
    \setlength{\tabcolsep}{1.0mm}
\resizebox{1.0\linewidth}{!}{
    \begin{tabular}{lcccccccccc}
    \toprule
Backbone & Vanilla & Mixup & CutMix & ManifoldMix & SaliencyMix & FMix  & PuzzleMix & Co-Mixup & AutoMix    & SAMix      \\ \hline
R-18     & 49.91   & 48.62 & 51.85  & 48.49       & 52.07       & 50.30 & 53.95     & 54.13    & 54.46      & \bf{57.08} \\
RX-50    & 53.38   & 50.27 & 57.16  & 49.73       & 58.21       & 59.80 & 59.34     & 59.76    & \bf{61.05} & 60.94      \\
    \bottomrule
    \end{tabular}
    }
    \label{tab:app_cub_wsol}
    \vspace{-0.5em}
\end{table*}

\subsection{Rules for Counting the Mixup Rankings}
\label{app:ranking}
We have summarized and analyzed a great number of mixup benchmarking results to compare and rank all the included mixup methods in terms of \textit{performance}, \textit{applicability}, and the \textit{overall} capacity. 
Specifically, regarding the \textit{performance}, we averaged the accuracy rankings of all mixup algorithms for each downstream task and averaged their robustness and calibration results rankings separately. Finally, these ranking results are averaged again to produce a comprehensive range of performance ranking results. 
As for the \textit{applicability}, we adopt a similar ranking computation scheme considering the \textit{time usage} and the \textit{generalizability} of the methods. 
With the \textit{overall} capacity ranking, we combined the performance and applicability rankings with a 1:1 weighting to obtain the final take-home rankings. 
For equivalent results, we take a tied ranking approach. For instance, if three methods are tied for first place, then the method that results in fourth place is recorded as second place by default.
Finally, we provide the comprehensive rankings as shown in Table~\ref{tab:method} and Table~\ref{tab:ranking}.

% \input{tables/tab_ranking}
% \twocolumn  % end onecolumn

%%%%%%%%%%%%%%%% rebuttal version 1 %%%%%%%%%%%%%%%%
\clearpage
% \renewcommand\thefigure{R\arabic{figure}}
% \renewcommand\thetable{R\arabic{table}}
% \setcounter{table}{0}
% \setcounter{figure}{0}
% \setcounter{page}{1}
% \nolinenumbers

\section{Reproduction Comparison}
\label{app:reproduction}
We provided the reproduction analysis of various mixup methods. Note that AutoMix~\citep{iclr2024adautomix}, SAMix~\citep{Li2021BoostingDV}, AdAutoMix~\citep{iclr2024adautomix}, and Decouple Mix~\citep{2022decouplemix} are \textbf{originally implemented in \texttt{OpenMixup}}, while the other popular mixup algorithms are reproduced based on their official source codes or descriptions. As shown in Table~\ref{tab:app_reproduction_cifar} and Table \ref{tab:app_reproduction_in1k}, the reproduced results are usually better than the original implementations because of the following reasons:
To ensure a fair comparison, we follow the standard training settings for various datasets. Without changing the training receipts, we applied the effective implementations of the basic training components. For example, we employ a better implementation of the cosine annealing learning rate scheduler (updating by iterations) instead of the basic version (updating by epochs). On CIFAR-100, we utilize the \texttt{RandomCrop} augmentation with a ``reflect" padding instead of the ``zero" padding. On Tiny-ImageNet, we utilize \texttt{RandomResizedCrop} with the cropping ratio of 0.2 instead of \texttt{RandomCrop} in some implementations.
On ImageNet-1K, we found that our reproduced results closely align with the reported performances, with any minor discrepancies (around $\pm$0.5\%) attributable to factors such as random initialization and specific hardware configurations.

\begin{table*}[ht]
    \vspace{-1.0em}
    \centering
    \caption{Comparison of benchmark results reproduced by \texttt{OpenMixup} and the official implementations on CIFAR-100 and Tiny-ImageNet. We report the top-1 accuracy and the training epoch.
    Note that AutoMix~\citep{iclr2024adautomix}, SAMix~\citep{Li2021BoostingDV}, AdAutoMix~\citep{iclr2024adautomix}, and Decouple Mix~\citep{2022decouplemix} are \textbf{originally implemented in \texttt{OpenMixup}}. The reproduced results are usually better than the original implementations because we applied the effective implementations of the standard training settings without changing the training receipts.
    }
    % \vspace{-4pt}
    \setlength{\tabcolsep}{1.5mm}
\resizebox{1.0\linewidth}{!}{
    \begin{tabular}{l|c|cc|cc}
    \toprule
Method                                                           & Publication & \multicolumn{2}{c|}{CIFAR-100 (R18)} & \multicolumn{2}{c}{Tiny-ImageNet (R18)} \\ \cline{3-6} 
                                                                 &             & \bf{Ours}           & Official       & \bf{Ours}             & Official        \\ \hline
\rowcolor[HTML]{FDF0E2} MixUp\citep{Zhang2018mixupBE}             & ICLR'2018   & \bf{79.24} (1200)   & 76.84 (1200)   & \bf{63.86} (400)      & 58.96 (400)     \\
\rowcolor[HTML]{FDF0E2} CutMix\citep{Yun2019CutMixRS}             & ICCV'2019   & \bf{78.29} (1200)   & 76.95 (1200)   & \bf{65.53} (400)      & 59.89 (400)     \\
\rowcolor[HTML]{FDF0E2} SmoothMix\citep{Lee2020SmoothMixAS}       & CVPRW'2020  & \bf{78.69} (800)    & 78.14 (800)    & \bf{66.65} (400)      & -               \\
\rowcolor[HTML]{FDF0E2} GridMix\citep{Baek2021GridMixSR}          & PR'2020     & \bf{78.72} (800)    & 78.09 (800)    & \bf{64.79} (400)      & 62.22 (400)     \\
\rowcolor[HTML]{FDF0E2} ResizeMix\citep{2020resizemix}            & CVMJ'2023   & \bf{79.19} (400)    & 79.05 (400)    & \bf{63.47} (400)      & 63.23 (400)     \\
\rowcolor[HTML]{FDF0E2} ManifoldMix\citep{Verma2019ManifoldMB}    & ICML'2019   & \bf{80.21} (1200)   & 79.98 (1200)   & \bf{64.15} (400)      & 60.24 (400)     \\
\rowcolor[HTML]{FDF0E2} FMix\citep{Harris2020FMixEM}              & arXiv'2020  & \bf{79.91} (400)    & 79.85 (400)    & \bf{63.47} (400)      & 61.43 (400)     \\
\rowcolor[HTML]{FDF0E2} AttentiveMix\citep{icassp2020Attentive}   & ICASSP'2020 & \bf{79.62} (200)    & 77.16 (200)    & \bf{64.01} (400)      & -               \\
\rowcolor[HTML]{FDF0E2} SaliencyMix\citep{Uddin2021SaliencyMixAS} & ICLR'2021   & \bf{79.75} (200)    & 76.11 (200)    & \bf{64.60} (400)      & -               \\
\rowcolor[HTML]{E7ECE4} PuzzleMix\citep{Kim2020PuzzleME}          & ICML'2020   & \bf{81.13} (800)    & 80.99 (800)    & \bf{65.81} (400)      & 63.48 (400)     \\
\rowcolor[HTML]{E7ECE4} AlignMixup\citep{2021alignmix}            & CVPR'2022   & \bf{82.27} (800)    & 82.12 (800)    & \bf{66.91} (400)      & 66.87 (400)     \\
    \bottomrule
    \end{tabular}
    }
    \label{tab:app_reproduction_cifar}
    \vspace{-0.5em}
\end{table*}

\begin{table*}[ht]
    \vspace{-1.0em}
    \centering
    \caption{Comparison of reproduced results with \texttt{OpenMixup} and the official implementations on ImageNet-1K. We report the top-1 accuracy and the training epoch.
    Our reproduced results closely align with the reported performances, with any minor discrepancies (around $\pm$0.5\%) attributable to factors such as random initialization and specific hardware configurations.
    }
    % \vspace{-4pt}
    \setlength{\tabcolsep}{1.8mm}
\resizebox{0.94\linewidth}{!}{
    \begin{tabular}{l|c|ccc}
    \toprule
Method                                                            & Publication & \multicolumn{3}{c}{ImageNet-1K}                \\
                                                                  &             & Backbone & \bf{Ours}        & Official         \\ \hline
\rowcolor[HTML]{FDF0E2} MixUp~\citep{Zhang2018mixupBE}             & ICLR'2018   & R50      & \bf{77.12} (100) & 77.01 (100)      \\
\rowcolor[HTML]{FDF0E2} CutMix~\citep{Yun2019CutMixRS}             & ICCV'2019   & R50      & \bf{77.17} (100) & 77.08 (100)      \\
\rowcolor[HTML]{FDF0E2} SmoothMix~\citep{Lee2020SmoothMixAS}       & CVPRW'2020  & R50      & \bf{77.84} (300) & 77.66 (300)      \\
\rowcolor[HTML]{FDF0E2} GridMix~\citep{Baek2021GridMixSR}          & PR'2020     & R50      & \bf{78.50} (300) & 78.25 (300)      \\
\rowcolor[HTML]{FDF0E2} ResizeMix~\citep{2020resizemix}            & CVMJ'2023   & R50      & \bf{78.91} (300) & 78.90 (300)      \\
\rowcolor[HTML]{FDF0E2} ManifoldMix~\citep{Verma2019ManifoldMB}    & ICML'2019   & R50      & \bf{77.01} (100) & 76.85 (100)      \\
\rowcolor[HTML]{FDF0E2} FMix~\citep{Harris2020FMixEM}              & arXiv'2020  & R50      & \bf{77.19} (100) & 77.03 (100)      \\
\rowcolor[HTML]{FDF0E2} AttentiveMix~\citep{icassp2020Attentive}   & ICASSP'2020 & DeiT-S   & \bf{80.32} (300) & 77.50 (300)      \\
\rowcolor[HTML]{FDF0E2} SaliencyMix~\citep{Uddin2021SaliencyMixAS} & ICLR'2021   & R50      & 78.46 (300)      & \bf{78.76} (300) \\
\rowcolor[HTML]{E7ECE4} PuzzleMix~\citep{Kim2020PuzzleME}          & ICML'2020   & R50      & \bf{77.54} (100) & 77.51 (100)      \\
\rowcolor[HTML]{E7ECE4} AlignMixup~\citep{2021alignmix}            & CVPR'2022   & R50      & 79.32 (300)      & \bf{79.50} (300) \\
\rowcolor[HTML]{CFEFFF} TransMix~\citep{cvpr2022transmix}          & CVPR'2022   & DeiT-S   & \bf{80.80} (300) & 80.70 (300)      \\
\rowcolor[HTML]{CFEFFF} SMMix~\citep{iccv2023SMMix}                & ICCV'2023   & DeiT-S   & \bf{81.10} (300) & \bf{81.10} (300) \\
    \bottomrule
    \end{tabular}
    }
    \label{tab:app_reproduction_in1k}
\end{table*}

\end{document}